\pdfoutput=1

\documentclass[11pt]{article}

\usepackage[final]{acl}

\usepackage{times}
\usepackage{latexsym}
\usepackage{amssymb} 
\usepackage{flushend}

\usepackage[T1]{fontenc}
\usepackage[ruled,vlined,linesnumbered]{algorithm2e}

\usepackage[utf8]{inputenc}

\usepackage{microtype}

\usepackage{inconsolata}

\usepackage{amsfonts}       
\usepackage{nicefrac}       

\usepackage[table]{xcolor}
\definecolor{lightgrayrow}{RGB}{245,245,245}

\newcommand{\uparrowgreen}{\textcolor{green!50!black}{$\uparrow$}}
\newcommand{\downarrowred}{\textcolor{red!70!black}{$\downarrow$}}

\usepackage{url}
\usepackage{tabularray}
\UseTblrLibrary{booktabs}
\usepackage{array}
\usepackage{tabularx}
\usepackage{graphicx}
\usepackage{tcolorbox}

\usepackage{booktabs}
\usepackage{multirow}
\usepackage{makecell}
\usepackage{bm}

\usepackage{amsmath}
\usepackage{multirow}
\usepackage{colortbl}
\usepackage{booktabs}
\renewcommand\arraystretch{0.95}
\usepackage{wrapfig}
\usepackage{xspace}
\usepackage{arydshln}
\usepackage{subcaption}
\usepackage{paralist}
\usepackage{enumitem}
\usepackage{hyperref}

\usepackage{tikz}
\definecolor{forestgrhl}{RGB}{84, 186, 111}
\definecolor{darkgrhl}{RGB}{136, 219, 158}
\definecolor{greenhl}{RGB}{201, 242, 212}
\definecolor{faintgrhl}{RGB}{244, 255, 241}
\definecolor{darkredhl}{RGB}{232, 173, 161}
\definecolor{redhl}{RGB}{242, 208, 201}
\definecolor{faintredhl}{RGB}{255, 244, 241}
\definecolor{whitehl}{RGB}{255, 255, 255}
\definecolor{yellowhl}{RGB}{255, 217, 164}
\definecolor{forestgreen}{rgb}{0.13, 0.55, 0.13}
\definecolor{bluehl}{RGB}{201, 225, 242}

\newcolumntype{B}[1]{>{\bfseries\raggedright\arraybackslash}p{#1}}
\newcolumntype{Z}{>{\raggedright\arraybackslash}X}
\tcbset{
seeexample/.style={
width=\columnwidth,
colback=whitehl,
colframe=forestgreen!70!black,
colbacktitle=greenhl!55!white,
coltitle=black,
fonttitle=\small\bfseries,
fontupper=\small,
boxrule=0.6pt,
arc=1mm,
left=4pt,
right=4pt,
top=4pt,
bottom=4pt,
toptitle=2pt,
bottomtitle=2pt
}
}

\title{SSE-Bio: A Structured Self-Evolving Agent with Agentic Retrieval Policy for Multi-Hop Biomedical Reasoning}

\author{
  Zhaohan Meng$^{1,2}$,
  Zaiqiao Meng$^{1,3}$,
  Siwei Liu$^{4}$,
  Hao Xu$^{2}$,
  Ke Yuan$^{5,6}$,
  Iadh Ounis$^{1}$   
  \vspace{2mm}\\
  $^{1}$School of Computing Science, University of Glasgow\\
  $^{2}$Brigham and Women's Hospital, Harvard Medical School, Harvard University\\
  $^{3}$Language Technology Lab, University of Cambridge\\
  $^{4}$School of Natural and Computing Science, University of Aberdeen\\
  $^{5}$School of Cancer Sciences, University of Glasgow\\
  $^{6}$Cancer Research UK Scotland Institute\\
  \texttt{z.meng.3@research.gla.ac.uk}\\
}

\begin{document}

\maketitle

\begin{abstract}
Biomedical multi-hop question answering (QA) requires models to connect evidence across intermediate entities such as diseases, drugs, proteins, and phenotypes. Existing agents typically rely on static retrieval workflows or coarse-grained prompt rewriting, which can lead to instruction drift when reasoning procedures need to be updated. We propose \textbf{SSE-Bio}, a structured self-evolving agent with an agentic retrieval policy for multi-hop biomedical reasoning. Instead of globally rewriting agent instructions, SSE-Bio maintains a structured state, selectively retrieves knowledge triplets and prior templates through a trainable proxy policy, and improves its reasoning memory through fine-grained template editing. To optimise retrieval decisions, we introduce a proxy-training strategy based on group relative policy optimization, where the proxy is improved through decision-contrastive groups over alternative retrieval choices. Experiments on three biomedical multi-hop QA benchmarks show that SSE-Bio consistently outperforms existing baselines, achieving an improvement of 6.56 absolute points over the strongest self-evolving baseline on BioHopR.\footnote{The source code for SSE-Bio is publicly available at  \href{https://github.com/ZhaohanM/SSE-Bio}{https://github.com/ZhaohanM/SSE-Bio}.}
\end{abstract}

\section{Introduction}

Biomedical multi-hop question answering (QA) task is important for evidence-grounded biomedical discovery, such as disease understanding and drug repurposing~\cite{xiang2025knowledge}. It requires models to answer questions by reasoning across multiple entities, such as diseases, drugs, and proteins, connected through complex biomedical relations~\cite{valizadeh2022ai,meng2024heterogeneous,wang2025trustworthy}. For example, a disease-oriented biomedical question may require a system to first identify the bridge protein associated with the disease and then infer the drug linked to that protein.  It is challenging because biomedical evidence is highly specialised and relationally structured, so errors in resolving the intermediate bridge entity can mislead the subsequent reasoning chain, while in multi-answer settings the system must avoid prematurely stopping after finding only a subset of valid answers~\cite{li2023leveraging,chen2025mdteamgpt,meng2026triplet}.

Recent retrieval-augmented generation (RAG) models have improved the performance of multi-hop biomedical QA by incorporating external knowledge into the reasoning process~\cite{he2024g,cao2024knowledge,rezaei-etal-2025-agentic}. However, without explicit verification and feedback over the reasoning path, they may leave intermediate bridge entities unresolved, causing the reasoning chain to drift and potentially returning incomplete answers in multi-answer scenarios. Interactive multi-agent systems make this process more flexible by enabling step-wise planning, execution, and feedback, while self-evolving agents further introduce long-term memory to adapt reasoning strategies across questions~\cite{srivastava2025debate,gao2025survey}.
Yet these agentic systems still face two limitations. First, they often rely on static retrieval workflows rather than explicitly learned retrieval control, which can lead to sub-optimal performance~\cite{huang2025biomni}. Second, their memory updates are often coarse-grained, typically through prompt-level rewriting, so that a local failure may alter the entire reasoning scaffold~\cite{jin2025stella,feng2026searl}. These limitations motivate a more controllable framework that combines explicit retrieval control with fine-grained memory evolution.

\begin{figure*}[htbp]
    \centering
    \includegraphics[width=\textwidth]{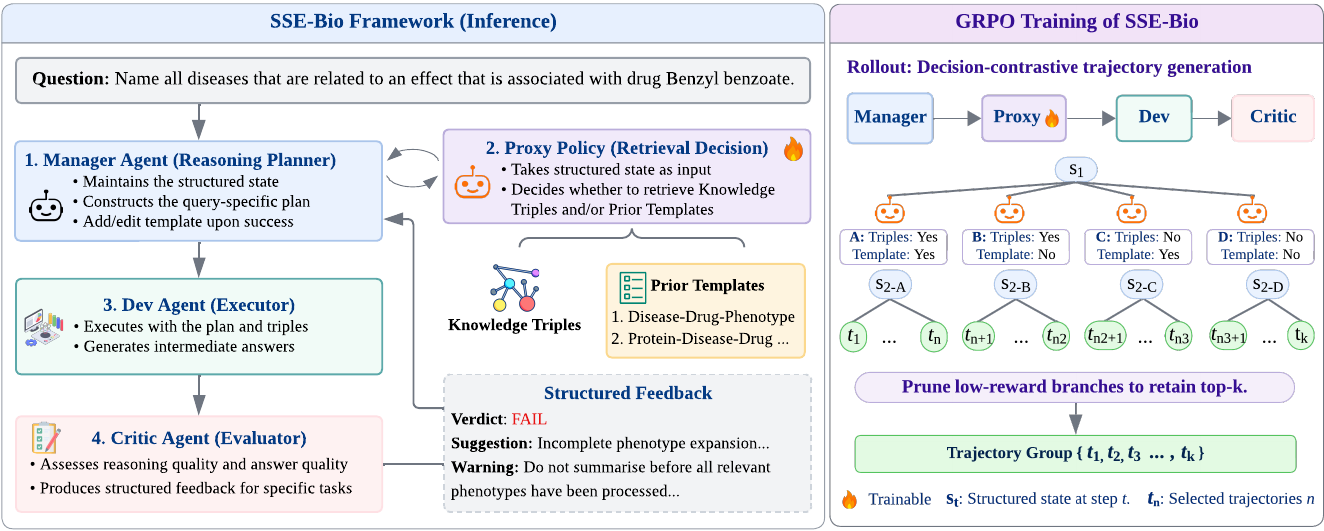}
    \caption{
        \textbf{Left:} The \textbf{Manager} maintains a structured state and template memory. Based on the current state, the \textbf{Proxy} decides whether to retrieve knowledge triplets and/or prior templates. The \textbf{Manager} then constructs a plan for the \textbf{Dev} agent, while the \textbf{Critic} returns structured feedback for iterative refinement. \textbf{Right:} Decision-contrastive trajectory generation explores alternative retrieval actions from the same state, and GRPO updates only the \textbf{Proxy} with an answer-grounded reward combining correctness and evidence support.}
        \label{fig:framework}
\end{figure*}

To address these limitations, we propose \textbf{SSE-Bio}, a structured self-evolving agent with an agentic retrieval policy for biomedical multi-hop QA. As shown in Fig.~\ref{fig:framework}, SSE-Bio combines structured reasoning memory with explicit retrieval management in a unified agentic framework. During inference, the \textbf{Manager} converts the structured state into a query-specific plan, which the \textbf{Dev} agent executes with the retrieved triplets, while the \textbf{Critic} provides structured feedback for iterative refinement. The \textbf{Manager} maintains two forms of memory: a structured state that captures the current reasoning status and serves as the agent's short-term memory, and a long-term memory of reusable templates distilled from prior successful cases. Crucially, instead of evolving memory through global rewriting, SSE-Bio updates templates through fine-grained editing, enabling local and auditable memory evolution that is designed to mitigate hallucination and instruction-drift risks from unconstrained template rewriting. In addition, SSE-Bio introduces a trainable \textbf{Proxy} model to manage retrieval explicitly by deciding whether knowledge triplets and/or prior templates should be retrieved at each reasoning step. Knowledge triplets are beneficial in the biomedical domain since they organize complex biomedical information into a clear structure, giving the Manager structured bridge entities and relation paths to inspect during multi-hop verification. In this way, SSE-Bio supports both dynamic evolution of reasoning memory over time and adaptive, selective retrieval from the current reasoning state, rather than relying on static retrieval pipelines or coarse-grained agent rewriting.

To optimise the retrieval behaviour of the \textbf{Proxy}, we adopt a two-stage training strategy. We first use supervised fine-tuning (SFT) to initialise the Proxy with retrieval decision labels that indicate whether \textit{knowledge triplets} and/or \textit{prior templates} should be retrieved under a given structured reasoning state. These labels are constructed by comparing alternative retrieval branches and selecting the decisions that lead to better downstream reasoning outcomes. We then further refine the Proxy with Group Relative Policy Optimization (GRPO)~\cite{shao2024deepseekmath}. In particular, we introduce \textit{decision-contrastive trajectory generation}, which expands alternative retrieval actions from the same structured reasoning state into contrastive rollout branches. The optimisation is driven by an answer-grounded reward that combines final answer correctness and evidence-supported reasoning behaviour, allowing GRPO to favour retrieval decisions that are both effective and well grounded. We evaluate SSE-Bio on three biomedical reasoning benchmarks, namely BioHopR, MedHop, and Humanity's Last Exam: Biomedicine, and show that it consistently outperforms strong baselines.
Our contributions are summarised as follows:
\begin{itemize}
    \item We propose \textbf{SSE-Bio}, a structured self-evolving agent for multi-hop biomedical reasoning, which unifies short-term structured state tracking, long-term template memory, and explicit retrieval management within a single agentic framework.
    \item We introduce a \textbf{fine-grained template evolution} mechanism that updates retrieved templates locally from successful trajectories, reducing unnecessary changes from the whole-template rewriting used in existing self-evolving agents.
    \item We design a \textbf{GRPO-based optimisation strategy} for the Proxy, which couples decision-contrastive trajectory generation with answer-grounded reward signals to compare retrieval branches under matched reasoning contexts and improve retrieval decisions through grounded supervision.
\end{itemize}

\section{Related Work}

\paragraph{Biomedical RAG Models.}
They improve biomedical QA by incorporating external knowledge into the reasoning process~\cite{tran2025rare,fang2025kirag}. In particular, i-MedRAG~\cite{xiong2024improving} strengthens biomedical QA by retrieving medical evidence to support answer generation. Similarly, MedGraphRAG~\cite{wu-etal-2025-medical} organises medical knowledge into graph structure so that retrieval can better reflect biomedical relations. Building on this direction, KRAGEN~\cite{matsumoto2024kragen} combines knowledge graphs with retrieval-augmented reasoning for biomedical problem solving. More recently, AMG-RAG~\cite{rezaei-etal-2025-agentic} further extends this line through agentic medical graph retrieval for evidence-grounded reasoning. Together, these methods show that external knowledge can improve biomedical multi-hop QA, but they still lack explicit verification and feedback over the evolving reasoning path, making them vulnerable to incorrect bridge-entity resolution and incomplete answers in multi-answer settings.

\paragraph{Biomedical Multi-agent Systems.}
They improve complex biomedical reasoning by enabling step-wise planning, execution, and feedback through specialised agent roles~\cite{almansoori2025self,chen2025mdteamgpt}. MDAgents~\cite{kim2024mdagents} improves medical reasoning through collaborative specialist agents. KGARevion~\cite{su2024kgarevion} introduces a knowledge-graph-based agent for knowledge-intensive biomedical QA. Biomni~\cite{huang2025biomni} further extends this direction through broader biomedical tool use. Beyond such multi-agent coordination, self-evolving systems aim to improve reasoning across questions by accumulating experience from previous trajectories. STELLA~\cite{jin2025stella} is a self-evolving biomedical agent that improves over time through an evolving Template Library and a dynamic Tool Ocean. Although it shows that long-term adaptation can improve biomedical reasoning, it provides limited control over when and what evidence should be retrieved, and its self-evolution updates templates by rewriting them from newly successful cases. In contrast, SSE-Bio explicitly couples state-aware retrieval control with structured long-term memory that evolves through fine-grained local editing, making both evidence use and memory evolution more bounded and controllable.

\section{Methodology}

We formulate SSE-Bio as an iterative agentic process in Sec.~\ref{sec:formulation}, then introduce its structured memory in Sec.~\ref{sec:structured_memory}, agentic retrieval policy in Sec.~\ref{sec:agentic_retrieval}, and \textbf{Proxy} policy training in Sec.~\ref{sec:proxy_training}.

\subsection{Agentic Formulation}
\label{sec:formulation}

We formulate biomedical multi-hop reasoning in SSE-Bio as an iterative agentic process involving four components: a \textbf{Manager}, a \textbf{Proxy}, a \textbf{Dev} agent, and a \textbf{Critic}. Given a biomedical question $q$, the goal of the agent is to resolve intermediate entities, verify the reasoning chain, and produce the answer. At reasoning round $t$, the \textbf{Manager} maintains a structured state $u_t$ as the short-term memory of the current question. Conditioned on $u_t$, the \textbf{Proxy} selects a retrieval action $a_t$, which determines whether external knowledge should be introduced at the current step. The selected evidence is returned to the \textbf{Manager}, which combines it with $u_t$ to construct a query-specific plan $\pi_t$ for the \textbf{Dev} agent. SSE-Bio also maintains a long-term memory $\mathcal{E}$ of reusable templates distilled from prior successful trajectories. 

The \textbf{Dev} agent executes the current plan and produces a reasoning trajectory $\tau_t$ and answer $\hat{y}_t$. The \textbf{Critic}, without access to the gold answer, assesses whether the trajectory and answer are coherent and sufficiently supported, and returns structured feedback $f_t$ for refinement. If the current reasoning fails, the \textbf{Manager} updates $u_t$ and replans the next round. If it succeeds, SSE-Bio updates $\mathcal{E}$ through fine-grained template editing or new-template distillation. Thus, SSE-Bio evolves through short-term structured state tracking within each question and long-term memory evolution across questions.

\subsection{Structured Memory}
\label{sec:structured_memory}


\paragraph{Short-term Memory.}
At each reasoning step, the \textbf{Manager} maintains a structured state $u_t$ as the short-term memory of the current question. Concretely, \textit{task type} denotes the relation pattern of the current question, \textit{information gap} specifies the missing evidence needed for the next reasoning step, \textit{feedback} records the Critic's structured comments on the current trajectory, and \textit{retrieval state} summarises what external evidence has already been retrieved. This state compactly represents the current reasoning state and makes retrieval decisions, planning, and refinement all grounded in the same explicit structure rather than scattered across free-form intermediate text. The structured state is updated across rounds to reflect reasoning progress and the structured feedback returned by the \textbf{Critic}. An example is provided in Fig.~\ref{fig:state_summary_example}.

\paragraph{Long-term Memory Evolution.}
Beyond the current question, SSE-Bio maintains a long-term memory of reusable templates distilled from prior successful cases. Each template stores reusable reasoning experience through a fixed set of fields: \textit{task type} identifies the question pattern to which the template applies, \textit{reasoning flow} records a reusable high-level solution procedure, \textit{verification criteria} specifies what conditions must be checked before returning an answer, \textit{tool-use policy} specifies what external tools can be used to support the reasoning process, and \textit{failure warning} highlights common mistakes to avoid. These templates are retrieved as planning priors for new questions. If the trajectory is aligned with a retrieved template, the \textbf{Manager} performs fine-grained template editing, updating only the local fields that require refinement. Otherwise, SSE-Bio performs new-template distillation, adding a new reusable template to memory. This design enables controllable memory evolution and constrains the kinds of broad template changes that can lead to instruction drift under coarse-grained rewriting. An example is provided in Fig.~\ref{fig:memory_edit_example}.

\subsection{Agentic Retrieval Policy}
\label{sec:agentic_retrieval}

The \textbf{Proxy} is responsible for explicitly controlling whether external knowledge should be introduced at each reasoning round. At reasoning round $t$, the Proxy selects a retrieval action:
\begin{equation}
a_t \sim \pi_{\theta}(\cdot \mid u_t), \qquad
a_t \in \mathcal{A} = 2^{\{\mathrm{KG},\mathrm{TPL}\}},
\end{equation}
where $\mathrm{KG}$ and $\mathrm{TPL}$ denote knowledge-triplet retrieval and template retrieval, respectively. This yields four actions: retrieving neither source, retrieving only knowledge triplets, retrieving only templates, or retrieving both. When $\mathrm{KG}\in a_t$, SSE-Bio uses the \textit{information gap} field in $u_t$ as the retrieval query, encodes it together with candidate biomedical triplets using BiCA~\cite{sinha2026bica}, and returns the top-$K$ triplets after schema and duplicate filtering. When $\mathrm{TPL}\in a_t$, SSE-Bio uses the \textit{task type} encoded in $u_t$ as the retrieval query, applies the same BiCA-based retrieval procedure to the long-term memory $\mathcal{E}$, and returns the top-$K$ compatible template. In this way, knowledge-triplet retrieval provides step-specific factual support, while template retrieval provides a reusable reasoning prior. Appendix~\ref{app:memory_examples} and~\ref{app:retrieval_protocol} report the memory schema and the retrieval budgets.

\subsection{Training of the Proxy Policy}
\label{sec:proxy_training}

The \textbf{Proxy} is the only trainable component in SSE-Bio; the Manager, Dev, Critic, retrievers, and reasoning environment remain fixed. The goal of training is to learn a retrieval policy $\pi_{\theta}(a_t \mid u_t)$ that makes better retrieval decisions under the current structured reasoning state.

\paragraph{SFT Initialisation.}
We first use SFT~\cite{wei2022finetuned,huang2025improve} to initialise the Proxy with retrieval decision pseudo-labels. For a given structured state $u_t$, we compare alternative retrieval branches and assign as pseudo-label the action whose rollout achieves the highest reward defined below. This stage provides a stable initialisation for the policy before reinforcement learning.

\paragraph{Decision Contrastive Trajectory Generation.}
We then refine the Proxy with GRPO~\cite{shao2024deepseekmath,park2025maporl}. SSE-Bio constructs \textit{decision-contrastive trajectory groups} by expanding alternative retrieval actions from the same reasoning state, forming matched rollout branches as shown in Fig.~\ref{fig:framework}. Branching is applied only in early reasoning rounds. Partial trajectories are pruned if they violate the required relation schema, leave the current information gap unresolved, or produce an unsupported bridge entity under the required schema; the remaining branches are ranked by intermediate answer-grounded reward signals, and only the top-$K$ are retained for further rollout. This keeps exploration focused while avoiding exponential growth. Appendix~\ref{app:training_details} provides the full pruning criteria and rollout-group statistics.

\paragraph{Reward Shaping.}
The Proxy is trained with an answer-grounded composite reward:
\begin{equation}
R(\tau)=
\lambda_o R_{\mathrm{out}}(\tau)+
\lambda_b R_{\mathrm{beh}}(\tau),
\end{equation}
where $\lambda_o+\lambda_b=1$.

\textbf{Outcome reward.}
The first term rewards final answer correctness:
\begin{equation}
R_{\mathrm{out}}(\tau)=\mathbf{1}\{\mathrm{corr}(\tau)\},
\end{equation}
where $\mathrm{corr}(\tau)$ indicates whether the final answer generated by trajectory $\tau$ matches the gold answer of the current training question.

\textbf{Behavioural reward.}
Final answer correctness alone is too sparse to distinguish trajectories with similar outcomes but different grounding quality. Therefore, we introduce a behavioural reward that measures whether key reasoning steps are supported by the retrieved evidence:
\begin{equation}
R_{\mathrm{beh}}(\tau)=
\frac{1}{|\mathcal{S}_{\tau}|}
\sum_{s\in\mathcal{S}_{\tau}}
\mathbf{1}\{s\ \text{is evidence-supported}\},
\end{equation}
where $\mathcal{S}_{\tau}$ denotes key steps such as bridge resolution and answer verification. A step is counted as supported only when it is grounded by a retrieved triplet or is consistent with the verification criteria specified by the retrieved template: bridge resolution requires a retrieved query--bridge relation, while answer verification requires retrieved or template-supported bridge--answer evidence. This term rewards grounded use of evidence rather than additional retrieval calls, with full support rules in Appendix table~\ref{tab:support_rules}.

\paragraph{Group-relative Optimisation.}
For a retained group $\{\tau_i\}_{i=1}^{K}$ with rewards $\{r_i\}_{i=1}^{K}$, we compute the group-relative advantage as follows:
\begin{equation}
A_i=
\frac{r_i-\mu(\mathbf{r})}
{\sigma(\mathbf{r})+\epsilon},
\end{equation}
where $\mathbf{r}$ denotes the rewards within the same trajectory group. Let $(u_i,a_i)$ denote the Proxy state and retrieval action associated with trajectory $\tau_i$. Since the Proxy outputs discrete retrieval actions rather than tokens, we apply GRPO at the action level and compare the current policy with the old sampling policy:
\begin{equation}
\rho_i(\theta)=
\frac{\pi_{\theta}(a_i\mid u_i)}
{\pi_{\theta_{\mathrm{old}}}(a_i\mid u_i)}.
\end{equation}
With $\bar{\rho}_i=\mathrm{clip}(\rho_i,1-\epsilon_c,1+\epsilon_c)$, the Proxy is trained by maximising the action-level GRPO objective:
\begin{equation}
\mathcal{J}_{\mathrm{GRPO}}
=
\mathbb{E}_{i}\!\left[
\min(\rho_i A_i,\bar{\rho}_i A_i)
-\beta_{\mathrm{KL}}D_{\mathrm{KL}}^i
\right],
\end{equation}
where $D_{\mathrm{KL}}^i$ regularises the retrieval-action distribution against the frozen SFT reference policy. Appendix~\ref{app:training_details} reports the pruning rules and rollout statistics used for this optimisation.

\begin{table*}[t]
\centering
\footnotesize
\setlength{\tabcolsep}{3.8pt}
\renewcommand{\arraystretch}{1.05}

\begin{tabular}{llcccccccc}
\toprule
\textbf{Family} & \textbf{Method}
& \multicolumn{4}{c}{\textbf{Single Answer}}
& \multicolumn{4}{c}{\textbf{Multi Answer}} \\
\cmidrule(lr){3-6}
\cmidrule(lr){7-10}
& & \textbf{Prec$_{H1}$}
& \textbf{Prec$_{H2}$}
& \textbf{Both$_{cor}$ \uparrowgreen}
& \textbf{Both$_{wr}$ \downarrowred}
& \textbf{Prec$_{H1}$}
& \textbf{Prec$_{H2}$}
& \textbf{Both$_{cor}$ \uparrowgreen}
& \textbf{Both$_{wr}$ \downarrowred} \\
\midrule

LLM & Llama-3.1-8B
& 0.13 & 0.04 & 0.00 & 99.83
& 0.07 & 0.01 & 0.00 & 99.92 \\

\rowcolor{gray!10}
LLM & Llama-3.1-70B
& 26.41 & 9.45 & 4.94 & 69.06
& 19.82 & 6.94 & 3.24 & 76.51 \\

LLM & GPT-4o
& 32.86 & 14.59 & 7.84 & 60.43
& 25.48 & 10.71 & 5.25 & 69.03 \\

\rowcolor{gray!10}
Medical LLM & HuatuoGPT-70B
& 0.15 & 0.00 & 0.00 & 99.84
& 0.09 & 0.00 & 0.00 & 99.91 \\

Medical LLM & HuatuoGPT-8B
& 0.21 & 0.03 & 0.00 & 99.76
& 0.11 & 0.01 & 0.00 & 99.88 \\

\rowcolor{gray!10}
Medical LLM & UltraMedical-8B
& 13.77 & 5.19 & 2.29 & 83.30
& 9.44 & 3.29 & 1.36 & 88.60 \\

RAG Model & i-MedRAG
& 25.20 & 7.48 & 4.43 & 71.75
& 16.86 & 5.73 & 2.49 & 79.90 \\

\rowcolor{gray!10}
RAG Model & AMG-RAG
& 28.13 & 8.94 & 5.61 & 68.47
& 19.26 & 6.62 & 3.38 & 77.21 \\

\midrule

Biomedical Agent & KGARevion
& 30.84 & 10.62 & 6.78 & 65.32
& 21.47 & 7.82 & 4.41 & 75.12 \\

\rowcolor{gray!10}
Biomedical Agent & DoctorAgent-RL
& 34.86 & 12.74 & 8.12 & 60.52
& 26.18 & 9.36 & 5.72 & 70.18 \\

Biomedical Agent & MedAgents
& 36.74 & 13.62 & 8.89 & 58.53
& 28.05 & 10.12 & 6.18 & 68.01 \\

\rowcolor{gray!10}
Biomedical Agent & Biomni
& 39.12 & 15.30 & 10.45 & 56.03
& 30.64 & 11.41 & 7.28 & 65.23 \\

Self-evolving Agent & STELLA
& 38.94 & 14.89 & 9.96 & 56.13
& 30.12 & 10.98 & 6.91 & 65.81 \\

\midrule
Self-evolving Agent & \textbf{SSE-Bio}
& \textbf{47.21$^{\dagger}$}
& \textbf{21.38$^{\dagger}$}
& \textbf{16.52$^{\dagger}$}
& \textbf{47.93$^{\dagger}$}
& \textbf{38.36$^{\dagger}$}
& \textbf{16.24$^{\dagger}$}
& \textbf{11.73$^{\dagger}$}
& \textbf{57.13$^{\dagger}$} \\

\bottomrule
\end{tabular}

\caption{Performance (\%) comparison on \textbf{BioHopR}. Prec$_{H1}$ and Prec$_{H2}$ evaluate the linked 1-hop and 2-hop questions, while Both$_{cor}$ and Both$_{wr}$ measure whether both questions in a pair are jointly correct or jointly wrong. $^{\dagger}$ indicates a significant improvement over STELLA (paired t-test, $p<0.05$).}
\label{tab:multihop_results}
\end{table*}

\begin{figure*}[t]
    \centering
    \includegraphics[width=\textwidth]{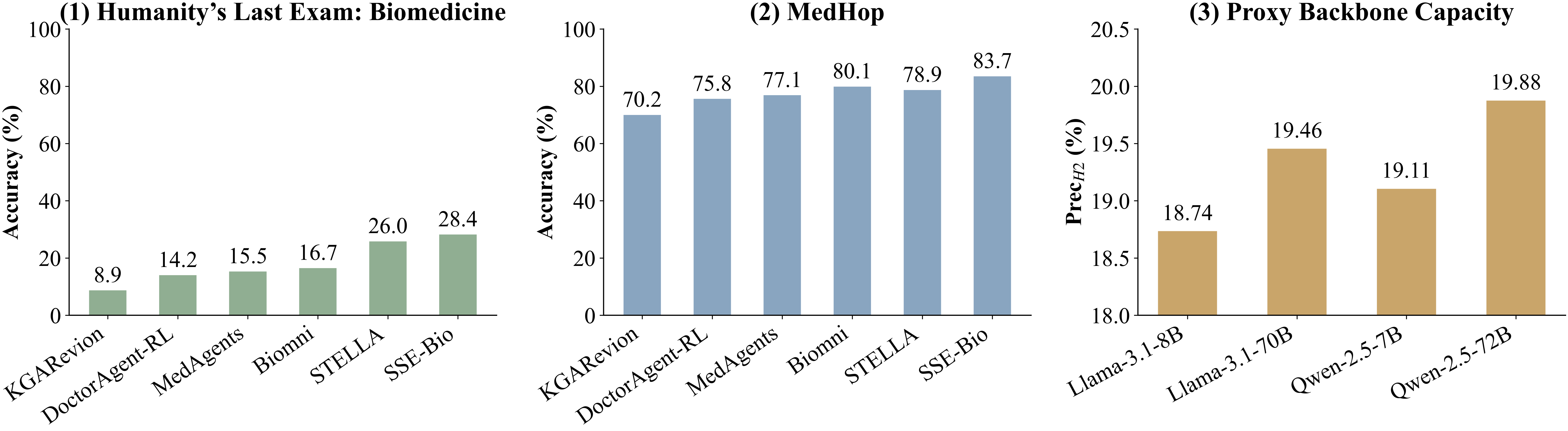}
    \caption{\textbf{(1)} Zero-shot accuracy (\%) comparison with agentic baselines on HLE. \textbf{(2)} Zero-shot accuracy (\%) comparison with agentic baselines on MedHop. \textbf{(3)} Performance (\%) of different proxy backbones on BioHopR.}
    \label{fig:see_bio_results}
\end{figure*}

\section{Experimental Setup}

\noindent\textbf{Datasets.}
We evaluate SSE-Bio on three biomedical reasoning benchmarks. BioHopR~\cite{kim2025biohopr} contains 7.63K paired instances constructed from complex biomedical relations, and we use a 7:3 train/test split, yielding approximately 5.34K training pairs and 2.29K test pairs. Each instance consists of a linked 1-hop and 2-hop question pair, and the benchmark supports both single-answer and multi-answer evaluation. We further assess cross-benchmark generalisation in a zero-shot setting on MedHop~\cite{welbl2018constructing} and Humanity's Last Exam: Biomedicine (HLE)~\cite{phan2025hle}. Appendix~\ref{app:datasets} provides further details.

\medskip
\noindent\textbf{Baselines.}
We compare SSE-Bio against representative baselines from five categories: general-purpose LLMs~\cite{grattafiori2024llama,openai2024gpt4o}, medical LLMs~\cite{chen2024huatuogpt}, retrieval-augmented generation~\cite{xiong2024improving,rezaei-etal-2025-agentic} methods, agent-based systems~\cite{su2024kgarevion,feng2026doctoragent,kim2024mdagents,huang2025biomni}, and self-evolving agents~\cite{jin2025stella}, which together cover non-agentic, retrieval-augmented, and self-evolving reasoning settings. For a fair comparison, all agentic baselines use the same backbone configuration as SSE-Bio: Gemini-2.5-Pro for Manager/Critic-style roles and Claude-4.5-Sonnet for execution. Appendix~\ref{app:baselines} provides details of each baseline.

\medskip
\noindent\textbf{Evaluation.}
For BioHopR, we follow the benchmark protocol and evaluate both single-answer and multi-answer instances using four metrics: \textbf{Prec$_{H1}$}, \textbf{Prec$_{H2}$}, \textbf{Both$_{cor}$}, and \textbf{Both$_{wr}$}. Prec$_{H1}$ and Prec$_{H2}$ measure precision on the linked 1-hop and 2-hop questions, while Both$_{cor}$ and Both$_{wr}$ measure the proportions of paired instances where both questions are answered correctly or incorrectly. Correctness is computed by embedding-based precision with cosine-similarity matching between predictions and gold answers. Higher values are better for Prec$_{H1}$, Prec$_{H2}$, and Both$_{cor}$, whereas lower values are better for Both$_{wr}$. For MedHop and HLE, we report zero-shot test accuracy. Appendix~\ref{app:evaluation_protocol} provides further details.

\medskip
\noindent\textbf{Implementation details.}
We fix Gemini-2.5-Pro as the Manager and Critic and Claude-4.5-Sonnet as the Dev agent, so performance differences can be attributed to retrieval-policy learning rather than changes in the reasoning backbones. SSE-Bio is trained on BioHopR and then transferred directly to MedHop and HLE for zero-shot evaluation. During evaluation, the template memory is frozen, and no instance from the BioHopR test split, MedHop, or HLE is used to create or edit templates. Additional details are provided in Appendix~\ref{app:implementation}.

\begin{table*}[t]
\centering
\footnotesize
\setlength{\tabcolsep}{3.2pt}
\renewcommand{\arraystretch}{1.02}

\scalebox{0.96}{%
\begin{tabular}{llrrrrrrrr}
\toprule
\textbf{Model / Setting} & \textbf{Backbone}
& \multicolumn{4}{c}{\textbf{Single Answer}}
& \multicolumn{4}{c}{\textbf{Multi Answer}} \\
\cmidrule(lr){3-6}
\cmidrule(lr){7-10}
& & \textbf{Prec$_{H1}$}
& \textbf{Prec$_{H2}$}
& \textbf{Both$_{cor}$ \uparrowgreen}
& \textbf{Both$_{wr}$ \downarrowred}
& \textbf{Prec$_{H1}$}
& \textbf{Prec$_{H2}$}
& \textbf{Both$_{cor}$ \uparrowgreen}
& \textbf{Both$_{wr}$ \downarrowred} \\
\midrule

\multicolumn{10}{c}{\textit{Effect of Structured State Tracking and Fine-grained Template Evolution}} \\
\midrule
\rowcolor{gray!10}
Free-form state + rewriting
& Qwen-72B
& 41.58 & 16.93 & 11.72 & 53.21
& 32.94 & 12.58 & 8.04 & 62.52 \\

Structured state + rewriting
& Qwen-72B
& 44.12 & 18.92 & 13.86 & 50.82
& 35.07 & 14.18 & 9.66 & 60.41 \\

\rowcolor{gray!10}
Free-form state + editing
& Qwen-72B
& 45.03 & 19.47 & 14.52 & 50.02
& 35.88 & 14.86 & 10.28 & 59.54 \\

Structured state + editing
& Qwen-72B
& \textbf{47.21}
& \textbf{21.38}
& \textbf{16.52}
& \textbf{47.93}
& \textbf{38.36}
& \textbf{16.24}
& \textbf{11.73}
& \textbf{57.13} \\

\midrule
\multicolumn{10}{c}{\textit{Component Ablation}} \\
\midrule
\rowcolor{gray!10}
w/o proxy
& Qwen-72B
& 39.74 & 15.88 & 10.62 & 55.00
& 30.81 & 11.92 & 7.05 & 64.32 \\

w/o triplets
& Qwen-72B
& 43.91 & 18.64 & 13.64 & 51.09
& 34.72 & 14.02 & 9.41 & 60.67 \\

\rowcolor{gray!10}
w/o prior template
& Qwen-72B
& 44.63 & 19.02 & 14.26 & 50.61
& 35.45 & 14.43 & 9.95 & 60.07 \\

\midrule
\multicolumn{10}{c}{\textit{Effect of Training and Boundary Analysis}} \\
\midrule
\rowcolor{gray!10}
SSE-Bio (SFT)
& Qwen-8B
& 44.67 & 19.11 & 14.10 & 50.32
& 35.68 & 14.52 & 9.84 & 59.64 \\

SSE-Bio (SFT)
& Qwen-72B
& 45.72 & 19.88 & 15.08 & 49.48
& 36.81 & 15.19 & 10.62 & 58.62 \\

\rowcolor{gray!10}
SSE-Bio (GRPO)
& Qwen-8B
& 46.08 & 20.64 & 15.62 & 48.90
& 37.24 & 15.71 & 11.08 & 58.13 \\

SSE-Bio (GRPO)
& Qwen-72B
& \textbf{47.21}
& \textbf{21.38}
& \textbf{16.52}
& \textbf{47.93}
& \textbf{38.36}
& \textbf{16.24}
& \textbf{11.73}
& \textbf{57.13} \\

\midrule
\rowcolor{gray!10}
SSE-Bio
& Qwen-72B
& 47.21 & 21.38 & 16.52 & 47.93
& 38.36 & 16.24 & 11.73 & 57.13 \\

+ Oracle Triplets
& Qwen-72B
& 47.86 & 22.31 & 17.34 & 47.17
& 39.02 & 17.04 & 12.36 & 56.30 \\

\rowcolor{gray!10}
+ Oracle Template
& Qwen-72B
& 47.63 & 21.98 & 17.02 & 47.41
& 38.75 & 16.81 & 12.14 & 56.58 \\

+ Oracle Triplets \& Template
& Qwen-72B
& \textbf{48.34}
& \textbf{22.94}
& \textbf{18.06}
& \textbf{46.78}
& \textbf{39.63}
& \textbf{17.64}
& \textbf{13.05}
& \textbf{55.78} \\

\bottomrule
\end{tabular}
}

\caption{Ablation study and boundary analysis on \textbf{BioHopR}.}
\label{tab:ablation_boundary}
\end{table*}

\begin{table*}[t]
\centering
\scriptsize
\setlength{\tabcolsep}{4.2pt}
\renewcommand{\arraystretch}{0.28}

\resizebox{\textwidth}{!}{%
\begin{tabular}{lcccccccc}
\toprule
\textbf{Relation Type}
& \multicolumn{4}{c}{\textbf{STELLA}}
& \multicolumn{4}{c}{\textbf{SSE-Bio}} \\
\cmidrule(lr){2-5}
\cmidrule(lr){6-9}
& \multicolumn{2}{c}{\textbf{Single Answer}}
& \multicolumn{2}{c}{\textbf{Multi Answer}}
& \multicolumn{2}{c}{\textbf{Single Answer}}
& \multicolumn{2}{c}{\textbf{Multi Answer}} \\
\cmidrule(lr){2-3}
\cmidrule(lr){4-5}
\cmidrule(lr){6-7}
\cmidrule(lr){8-9}
& \textbf{Prec$_{H1}$}
& \textbf{Prec$_{H2}$}
& \textbf{Prec$_{H1}$}
& \textbf{Prec$_{H2}$}
& \textbf{Prec$_{H1}$}
& \textbf{Prec$_{H2}$}
& \textbf{Prec$_{H1}$}
& \textbf{Prec$_{H2}$} \\
\midrule

\multicolumn{9}{c}{\textit{Same Query and Bridge}} \\
\midrule
\rowcolor{gray!10}
Disease:Drug:Phenotype
& 48.27 & 22.83 & 38.64 & 10.58
& 59.12$^{\dagger}$ & 34.26$^{\dagger}$ & 47.03$^{\dagger}$ & 16.82$^{\dagger}$ \\
Disease:Drug:Protein
& 46.75 & 10.36 & 37.42 & 4.37
& 52.18$^{\dagger}$ & 12.41$^{\dagger}$ & 41.02$^{\dagger}$ & 5.06$^{\dagger}$ \\
\rowcolor{gray!10}
Drug:Disease:Phenotype
& 50.18 & 20.67 & 33.15 & 8.74
& 58.74$^{\dagger}$ & 27.35$^{\dagger}$ & 37.91$^{\dagger}$ & 11.96$^{\dagger}$ \\
Drug:Disease:Protein
& 48.86 & 11.94 & 31.83 & 3.89
& 50.67$^{\dagger}$ & 13.18$^{\dagger}$ & 33.24$^{\dagger}$ & 4.52$^{\dagger}$ \\

\midrule
\multicolumn{9}{c}{\textit{Same Query and Target}} \\
\midrule
\rowcolor{gray!10}
Disease:Phenotype:Drug
& 27.74 & 16.13 & 15.68 & 6.47
& 30.12$^{\dagger}$ & 18.42$^{\dagger}$ & 17.08$^{\dagger}$ & 7.31$^{\dagger}$ \\
Disease:Protein:Drug
& 32.96 & 11.38 & 14.25 & 7.54
& 41.76$^{\dagger}$ & 18.93$^{\dagger}$ & 19.34$^{\dagger}$ & 12.84$^{\dagger}$ \\
\rowcolor{gray!10}
Drug:Phenotype:Disease
& 25.97 & 8.04 & 13.76 & 4.48
& 27.18$^{\dagger}$ & 9.07$^{\dagger}$ & 14.91$^{\dagger}$ & 5.18$^{\dagger}$ \\
Drug:Protein:Disease
& 30.43 & 17.35 & 20.87 & 7.49
& 38.84$^{\dagger}$ & 25.96$^{\dagger}$ & 26.72$^{\dagger}$ & 11.83$^{\dagger}$ \\

\midrule
\rowcolor{gray!10}
Phenotype:Disease:Drug
& 10.37 & 13.58 & 8.26 & 11.27
& 11.54$^{\dagger}$ & 15.06$^{\dagger}$ & 8.97$^{\dagger}$ & 15.47$^{\dagger}$ \\
Phenotype:Drug:Disease
& 23.95 & 6.54 & 14.78 & 4.96
& 31.86$^{\dagger}$ & 11.74$^{\dagger}$ & 19.52$^{\dagger}$ & 8.36$^{\dagger}$ \\
\rowcolor{gray!10}
Protein:Disease:Drug
& 34.28 & 7.46 & 15.84 & 4.17
& 36.11$^{\dagger}$ & 8.53$^{\dagger}$ & 16.92$^{\dagger}$ & 4.82$^{\dagger}$ \\
Protein:Drug:Disease
& 29.16 & 5.38 & 21.85 & 4.69
& 37.92$^{\dagger}$ & 9.86$^{\dagger}$ & 28.04$^{\dagger}$ & 7.74$^{\dagger}$ \\

\bottomrule

\end{tabular}
}

\caption{Relation-type breakdown on \textbf{BioHopR}. $^{\dagger}$ indicates a significant improvement (paired t-test, $p<0.05$).}
\label{tab:relation_breakdown}
\end{table*}

\section{Results and Analysis}

In this section, we answer five Research Questions (RQ) about SSE-Bio: overall performance (Sec.~\ref{sec:rq1}), the effect of the Proxy (Sec.~\ref{sec:rq2}), structured self-evolution (Sec.~\ref{sec:rq3}), retrieval headroom (Sec.~\ref{sec:rq4}), and robustness across biomedical relation types (Sec.~\ref{sec:rq5}).

\subsection{RQ1: How does SSE-Bio compare with baselines?}
\label{sec:rq1}

We compare SSE-Bio with strong baselines in Table~\ref{tab:multihop_results} and evaluate its cross-benchmark transfer in Fig.~\ref{fig:see_bio_results}. Table~\ref{tab:multihop_results} shows that SSE-Bio achieves the best performance across all four BioHopR metrics under both single-answer and multi-answer settings. In the single-answer setting, SSE-Bio significantly outperforms STELLA, on all four metrics (paired t-test, $p<0.05$), improving from 38.94 to 47.21 on Prec$_{H1}$, from 14.89 to 21.38 on Prec$_{H2}$, and from 9.96 to 16.52 on Both$_{cor}$, while reducing Both$_{wr}$ from 56.13 to 47.93. This shows that its gains extend beyond isolated hop-level prediction to paired multi-hop correctness. 

In the more challenging multi-answer setting, SSE-Bio again ranks first on all four metrics, reaching 38.36 on Prec$_{H1}$, 16.24 on Prec$_{H2}$, and 11.73 on Both$_{cor}$, showing that its advantage is preserved under the harder answer-coverage requirement. Compared with i-MedRAG and Biomni, SSE-Bio further improves single-answer Both$_{cor}$ by 12.09 and 6.07 points, indicating that state-aware retrieval and controllable memory evolution are more effective than fixed retrieval or generic biomedical agents. Fig.~\ref{fig:see_bio_results} further shows that SSE-Bio transfers strongly to MedHop and reaches 28.4\% accuracy on HLE. These results show that SSE-Bio consistently outperforms strong baselines while also exhibiting generalisable reasoning ability. Token, latency, and sensitivity analyses are in Appendices~\ref{efficiency} and~\ref{app:sensitivity}.

\subsection{RQ2: How does the Proxy affect SSE-Bio?}
\label{sec:rq2}

We evaluate whether the Proxy is necessary, whether it improves over fixed retrieval schedules, and whether GRPO further improves retrieval policy learning. In Table~\ref{tab:ablation_boundary}, the w/o Proxy setting, which retrieves neither knowledge triplets nor prior templates, reduces single-answer Both$_{cor}$ from 16.52 to 10.62 and increases Both$_{wr}$ from 47.93 to 55.00, indicating that retrieval control is important. Appendix~\ref{app:additional_analysis} further shows that the learned Proxy outperforms fixed retrieval schedules under the same backbone and retrieval budget, confirming that the gain is not simply due to retrieving more context. The training block in Table~\ref{tab:ablation_boundary} shows that GRPO improves over SFT. With the Qwen-72B Proxy, single-answer Both$_{cor}$ rises from 15.08 to 16.52 and Both$_{wr}$ falls from 49.48 to 47.93, with the same trend in the multi-answer setting. Appendix~\ref{app:additional_analysis} further shows that larger Proxy backbones yield stronger performance. Overall, the Proxy acts as an effective trainable control point for state-dependent retrieval decisions.

\subsection{RQ3: What are the effects of structured self-evolution?}
\label{sec:rq3}

We evaluate structured self-evolution by comparing free-form and structured variants of short-term state tracking, together with coarse-grained rewriting and fine-grained template editing for long-term memory evolution, in Table~\ref{tab:ablation_boundary}. Replacing both structured state and fine-grained editing with free-form state and coarse-grained rewriting reduces single-answer Both$_{cor}$ from 16.52 to 11.72 and increases Both$_{wr}$ from 47.93 to 53.21. Introducing only structured state or only fine-grained editing recovers part of this gap, yielding Both$_{cor}$ values of 13.86 and 14.52, respectively, but neither matches the full structured design. This pattern suggests that structured state tracking helps maintain an explicit reasoning status for retrieval and refinement, while fine-grained template editing updates reusable memory locally rather than rewriting the whole reasoning scaffold. Together, they provide a more controlled form of self-evolution. Appendix~\ref{app:additional_analysis} further supports this interpretation by showing that fine-grained editing changes fewer unrelated fields and introduces fewer unsupported or contradictory instructions than coarse rewriting.

\subsection{RQ4: How much headroom remains under stronger retrieval?}
\label{sec:rq4}

We evaluate the remaining headroom of SSE-Bio from two perspectives in Table~\ref{tab:ablation_boundary}: the \textit{Component Ablation} block tests whether triplets and prior templates each contribute to performance, and the \textit{Boundary Analysis} block estimates how much further the system could improve with stronger evidence and template selection. In the component ablation block, removing triplets or prior templates reduces single-answer Both$_{cor}$ from 16.52 to 13.64 and 14.26, respectively, showing that factual evidence and reusable reasoning priors are complementary rather than redundant. In the boundary analysis block of Table~\ref{tab:ablation_boundary}, oracle triplets improve single-answer Both$_{cor}$ to 17.34, oracle templates improve it to 17.02, and combining both reaches 18.06 while reducing Both$_{wr}$ to 46.78. The multi-answer setting follows the same pattern, with combined oracle retrieval increasing Both$_{cor}$ from 11.73 to 13.05 and reducing Both$_{wr}$ from 57.13 to 55.78. These results show that SSE-Bio already learns useful retrieval behaviour, but still has measurable headroom under stronger evidence and template retrieval. Appendix~\ref{app:retrieval_robustness} further evaluates robustness under injected noisy triplets.

\subsection{RQ5: How does SSE-Bio perform across different relation types?}
\label{sec:rq5}

To assess whether SSE-Bio's gains generalise across biomedical reasoning patterns, we report a relation-type breakdown on BioHopR and compare SSE-Bio with STELLA. As shown in Table~\ref{tab:relation_breakdown}, SSE-Bio improves over STELLA across all listed relation types, suggesting that its advantage is not confined to a specific query or bridge configuration. The gains are especially clear for relations involving stronger second-hop reasoning. For example, Disease:Drug:Phenotype improves from 22.83 to 34.26 on single-answer Prec$_{H2}$ and from 10.58 to 16.82 on multi-answer Prec$_{H2}$. By contrast, Drug:Disease:Protein shows more modest gains, from 11.94 to 13.18 and from 3.89 to 4.52, indicating that some relation structures remain challenging. Multi-answer scores are consistently lower than single-answer scores, reflecting the stricter requirement of retrieving multiple correct answers. Overall, SSE-Bio provides consistent relation-level improvements, with larger gains on relation types requiring effective bridge tracking and second-hop reasoning. Appendix~\ref{app:stella_failure_case} provides a concrete failure case showing how relation-specific retrieval and local verification-field editing correct an unsupported bridge-entity error.

\section{Conclusions}\label{sec:conclusion}

We presented SSE-Bio, a structured self-evolving agent for biomedical multi-hop reasoning that combines state-aware retrieval control through a trainable proxy with controllable memory evolution through structured templates and fine-grained editing. This design enables adaptive retrieval and reasoning without fixed retrieval pipelines or unconstrained self-rewriting. Experiments on three benchmarks show that SSE-Bio consistently outperforms strong baselines. On BioHopR, SSE-Bio improves over the self-evolving baseline, STELLA, by 6.56 points on single-answer Both$_{cor}$, while also achieving the best performance under multi-answer settings. Ablation and relation-type analyses further show that these gains are supported by effective retrieval control and structured self-evolution. Overall, our results suggest that combining structured retrieval control with fine-grained self-evolution is a promising step towards more controllable biomedical reasoning agents.

\section*{Limitations}

We identify the following limitations of our work. (1) SSE-Bio relies on retrieved biomedical triplets and prior templates to support multi-hop reasoning. Although our retrieval design is effective, the oracle analysis shows that stronger evidence and template matching can still improve downstream performance. In future work, we plan to explore more advanced biomedical retrieval and template matching strategies. (2) SSE-Bio depends on long-term template memory to support structured self-evolution. While fine-grained editing makes memory updates more controllable than unconstrained rewriting, memory growth may still introduce redundant or outdated templates over time. Future work should therefore investigate memory consolidation, pruning, and provenance tracking. (3) Our proxy analysis shows that stronger proxy backbones consistently improve retrieval control. Although SSE-Bio already performs well with the current proxy settings, it remains an open question how far proxy scaling can further improve performance. We leave the exploration of larger and more efficient proxy architectures to future work. (4) SSE-Bio trains only the Proxy while relying on fixed commercial Manager, Dev, and Critic agents, so reproduction may be affected by prompt details and API behaviour. Moreover, SFT labels and GRPO rewards are induced from rollouts within the same fixed agent environment; future work should test whether the learned retrieval policy transfers across different agent backbones.

\section*{Acknowledgements}
The authors acknowledge the use of resources provided by the Isambard-AI National AI Research Resource (AIRR)~\cite{mcintosh2024isambard}. Isambard-AI is operated by the University of Bristol and is funded by the UK Government's Department for Science, Innovation and Technology (DSIT) via UK Research and Innovation; and the Science and Technology Facilities Council [ST/AIRR/I-A-I/1023].

K.Y. acknowledges support from Cancer Research UK (EDDPGM-Nov21/100001, DRCMDP-Nov23/100010 and core funding to the CRUK Scotland Institute (A31287)), BBSRC BB/V016067/1, Prostate Cancer UK MA-TIA22-001 and EU Horizon 2020 grant ID: 101016851.

\bibliography{custom}

\appendix
\clearpage

\section{Appendix}\label{appendix}

\subsection{Computing Hardware}\label{model_details}

All experiments were conducted on a compute node equipped with six \textbf{NVIDIA GH200 Grace Hopper superchips}. This hardware provided sufficient memory and computational capacity for the training and inference settings used in this work.

\subsection{Examples of Structured Memory}
\label{app:memory_examples}

To make the structured memory design in Sec.~\ref{sec:structured_memory} more concrete, we provide the field schema and two illustrative examples. Table~\ref{tab:memory_schema} defines the fields used in SSE-Bio's short-term structured state and long-term template memory. The short-term state captures the current reasoning status for retrieval control and replanning, while the long-term template stores reusable reasoning priors that can be retrieved and locally edited for future questions. Figures~\ref{fig:state_summary_example} and~\ref{fig:memory_edit_example} instantiate these definitions with concrete examples of a structured state and a fine-grained template update.

Figure~\ref{fig:state_summary_example} shows a representative structured state used as short-term memory during iterative reasoning. It illustrates how the \textbf{Manager} organises the task type, information gap, Critic feedback, and retrieval state into a compact representation for retrieval control and next-step planning.

Figure~\ref{fig:memory_edit_example} shows an example of long-term template memory and its fine-grained update after a successful trajectory. The example highlights how SSE-Bio keeps the task type fixed while locally editing fields such as the reasoning flow, verification criteria, tool-use policy, and failure warning, rather than rewriting the entire template.

\begin{figure}[t]
\centering
\begin{tcolorbox}[seeexample,title={Example: structured state}]
\begingroup
\renewcommand{\arraystretch}{1.08}
\begin{tabularx}{\linewidth}{@{}B{0.35\linewidth}Z@{}}
Task type & Disease--Protein--Drug (multi answer) \\
Information gap & The disease is known, but the bridge proteins and the candidate drugs must still be verified exhaustively. \\
Feedback & The Critic requires explicit disease--protein and protein--drug support before returning the final drug set. \\
Retrieval state & Knowledge triplets retrieved for disease--protein; one compatible template retrieved for multi-step verification. \\
\end{tabularx}
\endgroup
\end{tcolorbox}
\caption{Example of the structured state used as short-term memory for retrieval control and planning.}
\label{fig:state_summary_example}
\end{figure}

\begin{figure}[!ht]
\centering
\begin{tcolorbox}[seeexample,title={Example: Long-term Template Memory and Fine-grained Update}]
\begingroup
\renewcommand{\arraystretch}{1.08}
\begin{tabularx}{\linewidth}{@{}B{0.34\linewidth}Z@{}}
Task type & Disease--Protein--Drug (multi answer) \\
Reasoning flow & Identify candidate proteins associated with the disease; retrieve candidate drugs linked to those proteins... \\
Verification criteria & Ensure each predicted drug is linked to at least one protein associated with the disease... \\
Tool-use policy & Python for aggregation or deduplication... \\
\rowcolor{gray!10}
Failure warning (before) & Avoid returning a partial drug set after verifying only one plausible disease-associated protein. \\
\rowcolor{bluehl}
Failure warning (after) & Avoid finalising the answer before checking whether alternative disease-associated proteins change the verified drug set. \\
\end{tabularx}
\endgroup
\end{tcolorbox}
\caption{Example of long-term template memory and its fine-grained update after a successful case. The grey row shows the original field implicated by the failure, and the blue row shows the updated field. Other fields remain unchanged.}
\label{fig:memory_edit_example}
\end{figure}

\begin{table}[!ht]
\centering
\small
\begin{tabularx}{\linewidth}{@{}l l X@{}}
\toprule
\textbf{Memory} & \textbf{Field} & \textbf{Role} \\
\midrule
State $u_t$ & Task type & template retrieval query \\
State $u_t$ & Information gap & triplet retrieval query \\
State $u_t$ & Feedback & Critic guidance for refinement \\
State $u_t$ & Retrieval state & retrieved evidence summary \\
\midrule
Template $x$ & Task type & template retrieval key \\
Template $x$ & Reasoning flow & reusable reasoning steps \\
Template $x$ & Verification criteria & answer-checking requirements \\
Template $x$ & Tool-use policy & external tools available for supporting the reasoning process \\
Template $x$ & Failure warning & common failure pattern \\
\bottomrule
\end{tabularx}
\caption{Field schema of the structured state and template memory in SSE-Bio.}
\label{tab:memory_schema}
\end{table}

\subsection{Prompt and API Sensitivity}
\label{app:sensitivity}

To assess whether SSE-Bio's gains depend on brittle prompt wording or a specific commercial API endpoint, we conduct two controlled sensitivity analyses on the BioHopR test split.

\paragraph{Prompt sensitivity.}
We vary the system prompts of the Manager, Dev, and Critic agents across three settings: no system prompt (\textit{Empty}), a generic prompt (``You are a helpful biomedical assistant.'', \textit{Basic}), and the role-specific prompts used in SSE-Bio. All other components, including the user prompts, structured input and output schemas, retrieval action space, template format, stopping criterion, answer format, model APIs, and trained Proxy, are kept fixed. As shown in Table~\ref{tab:prompt_sensitivity}, replacing the role-specific system prompts with Empty or Basic prompts causes only small drops, suggesting that SSE-Bio is not driven by brittle system-prompt wording.

\begin{table*}[t]
\centering
\small
\begin{tabular*}{\textwidth}{@{\extracolsep{\fill}}lcccccccc@{}}
\toprule
\textbf{System Prompt}
& \multicolumn{4}{c}{\textbf{Single Answer}}
& \multicolumn{4}{c}{\textbf{Multi Answer}} \\
\cmidrule(lr){2-5}\cmidrule(lr){6-9}
& \textbf{Prec$_{H1}$} & \textbf{Prec$_{H2}$} & \textbf{Both$_{cor}$} & \textbf{Both$_{wr}$}
& \textbf{Prec$_{H1}$} & \textbf{Prec$_{H2}$} & \textbf{Both$_{cor}$} & \textbf{Both$_{wr}$} \\
\midrule
Empty   & 46.38 & 20.71 & 16.08 & 48.62 & 37.64 & 15.61 & 11.21 & 58.04 \\
Basic   & 46.72 & 20.96 & 16.27 & 48.31 & 37.91 & 15.87 & 11.44 & 57.68 \\
SSE-Bio & \textbf{47.21} & \textbf{21.38} & \textbf{16.52} & \textbf{47.93}
        & \textbf{38.36} & \textbf{16.24} & \textbf{11.73} & \textbf{57.13} \\
\bottomrule
\end{tabular*}
\caption{Prompt-sensitivity analysis on BioHopR. Replacing role-specific system prompts causes only small drops, indicating that SSE-Bio is not driven by brittle prompt wording.}
\label{tab:prompt_sensitivity}
\end{table*}

\paragraph{API sensitivity.}
We replace the Manager and Critic backbones while keeping the Dev agent, trained Proxy, retriever, template memory, prompts, and evaluation protocol fixed. As shown in Table~\ref{tab:api_sensitivity}, alternative Manager/Critic APIs lead to moderate but not catastrophic degradation.

\begin{table*}[t]
\centering
\small
\begin{tabular*}{\textwidth}{@{\extracolsep{\fill}}llcccccccc@{}}
\toprule
\textbf{Manager / Critic} & \textbf{Dev}
& \multicolumn{4}{c}{\textbf{Single Answer}}
& \multicolumn{4}{c}{\textbf{Multi Answer}} \\
\cmidrule(lr){3-6}\cmidrule(lr){7-10}
& & \textbf{Prec$_{H1}$} & \textbf{Prec$_{H2}$} & \textbf{Both$_{cor}$} & \textbf{Both$_{wr}$}
& \textbf{Prec$_{H1}$} & \textbf{Prec$_{H2}$} & \textbf{Both$_{cor}$} & \textbf{Both$_{wr}$} \\
\midrule
Gemini-2.5-Pro   & Claude-4.5-Sonnet & \textbf{47.21} & \textbf{21.38} & \textbf{16.52} & \textbf{47.93} & \textbf{38.36} & \textbf{16.24} & \textbf{11.73} & \textbf{57.13} \\
Gemini-2.5-Flash & Claude-4.5-Sonnet & 45.96 & 20.14 & 15.86 & 49.08 & 37.02 & 15.12 & 11.04 & 58.36 \\
GPT-5.5          & Claude-4.5-Sonnet & 46.54 & 20.83 & 16.11 & 48.57 & 37.68 & 15.74 & 11.38 & 57.91 \\
GPT-5.5-Mini     & Claude-4.5-Sonnet & 45.31 & 19.62 & 15.42 & 49.74 & 36.44 & 14.83 & 10.72 & 58.92 \\
\bottomrule
\end{tabular*}
\caption{API-sensitivity analysis on BioHopR. Alternative Manager/Critic APIs lead to moderate but not catastrophic degradation.}
\label{tab:api_sensitivity}
\end{table*}

Together, these analyses indicate that SSE-Bio is reasonably stable under both prompt and API variations.

\subsection{Efficiency Details}
\label{efficiency}

Table~\ref{tab:efficiency} compares SSE-Bio with the strongest agentic baselines on the BioHopR test split in terms of total token usage and single-answer Both$_{cor}$. SSE-Bio achieves the best paired-correctness performance, reaching 16.52 on Both$_{cor}$, compared with 10.45 for Biomni and 9.96 for STELLA. This gain, however, comes with higher token usage: SSE-Bio uses 28.3M tokens, compared with 23.6M for Biomni and 21.2M for STELLA. These results indicate that SSE-Bio improves effectiveness substantially, while incurring additional token cost relative to the compared baselines.

\begin{table}[!ht]
\centering
\small
\renewcommand{\arraystretch}{1.1}
\begin{tabular*}{\columnwidth}{@{\extracolsep{\fill}}lrr@{}}
\toprule
\textbf{Model} & \textbf{Tokens (M)} & \textbf{Both$_{cor}$} \\
\midrule
Biomni  & 23.6 & 10.45 \\
STELLA  & 21.2 & 9.96 \\
SSE-Bio & 28.3 & 16.52 \\
\bottomrule
\end{tabular*}
\caption{Efficiency comparison on the BioHopR test split.}
\label{tab:efficiency}
\end{table}

Beyond token usage, Table~\ref{tab:inference_cost} reports end-to-end per-example inference cost, including all Manager, Dev, Critic, retrieval, and local Proxy operations, and Table~\ref{tab:training_cost} reports the one-time Proxy training cost.

\begin{table}[!ht]
\centering
\small
\begin{tabular*}{\columnwidth}{@{\extracolsep{\fill}}lrrrr@{}}
\toprule
\textbf{Method} & \makecell{\textbf{Tokens}\\\textbf{/ Ex.}} & \makecell{\textbf{LLM Calls}\\\textbf{/ Ex.}} & \makecell{\textbf{Latency}\\\textbf{/ Ex.}} & \makecell{\textbf{API Cost}\\\textbf{/ Ex.}} \\
\midrule
Biomni  & 10.3K & 5.8 & 18.4s & \$0.039 \\
STELLA  & 9.3K  & 4.9 & 16.7s & \$0.035 \\
SSE-Bio & 12.4K & 6.7 & 22.1s & \$0.047 \\
\bottomrule
\end{tabular*}
\caption{Per-example inference cost on BioHopR. SSE-Bio adds moderate overhead relative to STELLA (3.1K tokens, 1.8 calls, 5.4s, and approximately \$0.012 per example) while substantially improving paired correctness. The local Proxy adds no commercial API calls; its latency is included in the end-to-end measurement.}
\label{tab:inference_cost}
\end{table}

\begin{table}[!ht]
\centering
\small
\begin{tabular*}{\columnwidth}{@{\extracolsep{\fill}}lrrr@{}}
\toprule
\textbf{Stage} & \textbf{Tokens} & \textbf{Data / Rollouts} & \textbf{GPU-h} \\
\midrule
Proxy SFT  & 41.6M  & 12.6K labelled pairs & 42 \\
Proxy GRPO & 132.4M & 38.5K rollouts       & 96 \\
\bottomrule
\end{tabular*}
\caption{One-time Proxy training cost (138 GPU-hours in total), not incurred during test-time inference.}
\label{tab:training_cost}
\end{table}

\begin{figure}[!ht]
\centering
\includegraphics[width=\linewidth]{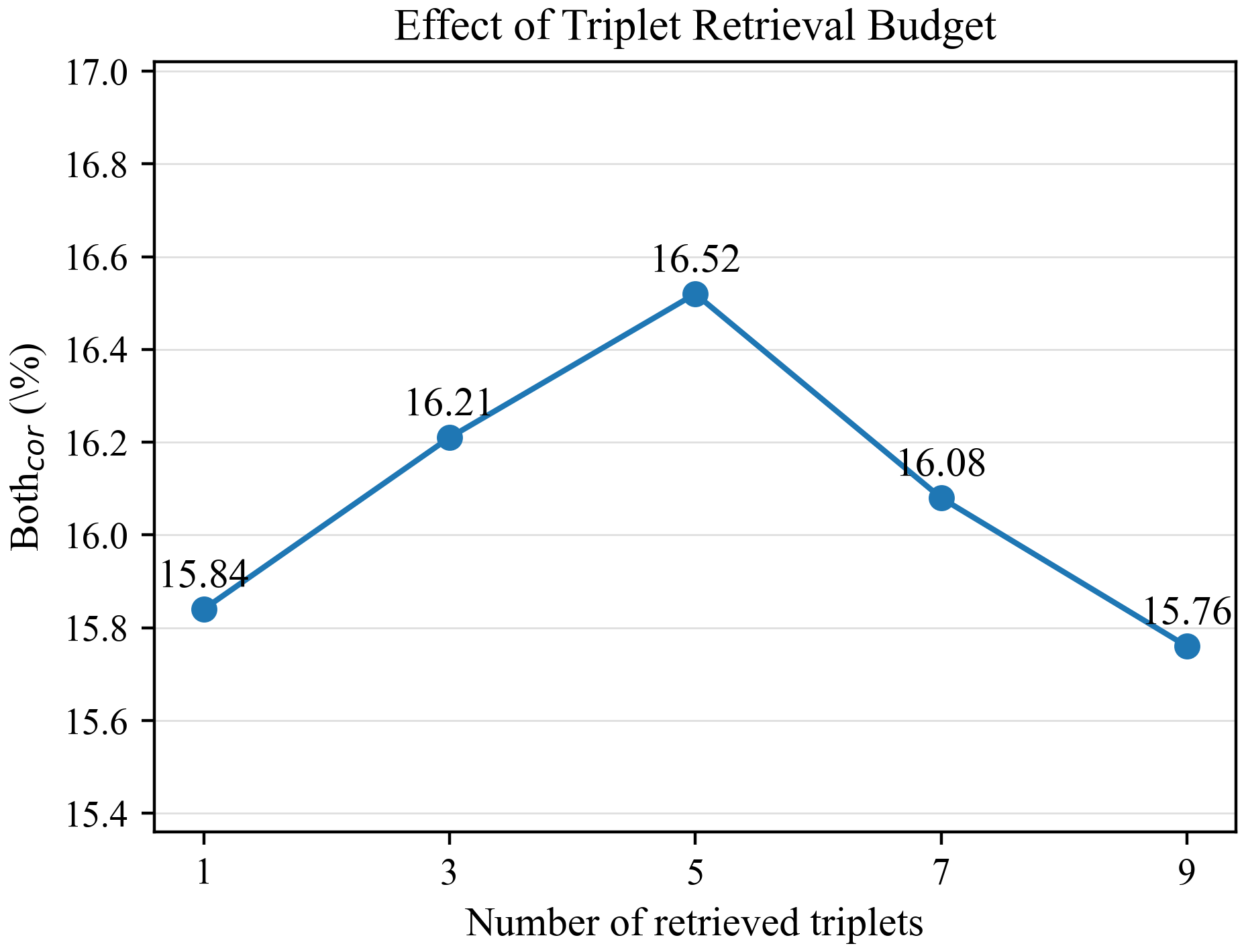}
\caption{Effect of the triplet retrieval budget on BioHopR.}
\label{fig:retrieval_budget_triplets}
\end{figure}

\begin{figure}[!ht]
\centering
\includegraphics[width=\linewidth]{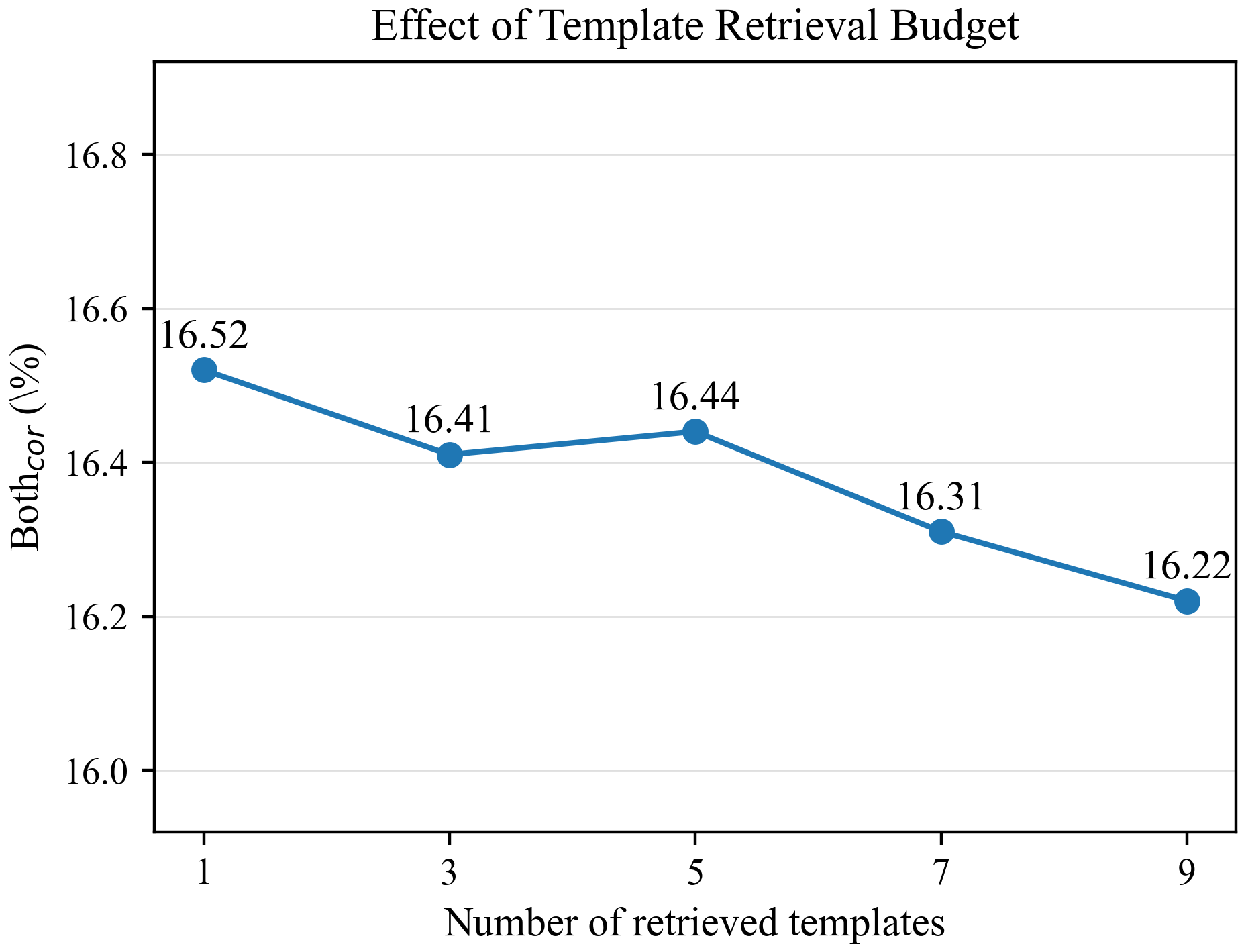}
\caption{Effect of the template retrieval budget on BioHopR.}
\label{fig:retrieval_budget_templates}
\end{figure}

\subsection{Implementation Details}
\label{app:implementation}
SSE-Bio uses a four-agent architecture consisting of a Manager, a Proxy, a Dev agent, and a Critic. The Manager and Critic are implemented with Gemini-2.5-Pro, and the Dev agent uses Claude-4.5-Sonnet. These fixed commercial agents execute candidate rollouts and provide critic feedback for both SFT data construction and GRPO trajectory groups. The Proxy is the only trainable component and is instantiated with open-weight backbones, including Llama-3.1-8B/70B and Qwen-2.5-7B/72B. For compactness, result tables abbreviate Qwen-2.5-7B/72B as Qwen-8B/72B. We perform full-parameter fine-tuning for all Proxy backbones, including Qwen-2.5-72B, using the hardware described in Appendix~\ref{model_details}. Unless otherwise stated, reported SSE-Bio results use the Qwen-2.5-72B Proxy.

\paragraph{Potential risks and intended use.} Although SSE-Bio is evaluated only on benchmark biomedical reasoning tasks, incorrect answers or unsupported reasoning could be harmful if the system were used directly in biomedical or clinical decision-making. SSE-Bio should therefore be treated as a research system for evidence-grounded reasoning evaluation rather than as a tool for clinical diagnosis, treatment recommendation, or patient-facing deployment.

\paragraph{Artifact licences and intended use.}
We use publicly available datasets and models in accordance with their stated research-use terms and licences. BioHopR, MedHop, Humanity's Last Exam: Biomedicine, BiCA, Llama, Qwen, and the compared baseline systems are used only for research evaluation. Commercial LLM APIs, including Gemini-2.5-Pro and Claude-4.5-Sonnet, are used according to their provider terms of service. The released SSE-Bio code will be distributed for research use.

\begin{table}[t]
\centering
\footnotesize
\begin{tabular}{@{}ll@{}}
\toprule
\textbf{Hyperparameter} & \textbf{Value} \\
\midrule
Proxy backbone & Qwen-2.5-72B \\
SFT learning rate & $1\times10^{-5}$ \\
SFT batch size & 64 \\
SFT max length & 2048 \\
SFT \# epochs & 1 \\
GRPO learning rate & $5\times10^{-7}$ \\
GRPO KL coefficient & 0.002 \\
GRPO batch size & 1024 \\
GRPO max length & 2048 \\
GRPO LR scheduler & Cosine \\
Reward weights $(\lambda_o,\lambda_b)$ & $(0.8,\,0.2)$ \\
Retained trajectory cap $K$ & 8 \\
Triplet retrieval budget $k_{\mathrm{KG}}$ & 5 \\
Template retrieval budget $k_{\mathrm{TPL}}$ & 1 \\
Template edit threshold $\theta_{\mathrm{tpl}}$ & 0.75 \\
BioHopR matching threshold $\tau$ & 0.9 \\
\bottomrule
\end{tabular}
\caption{Key training and retrieval hyperparameters of SSE-Bio.}
\label{tab:hyperparams}
\end{table}

\subsection{Agentic Retrieval}
\label{app:retrieval_protocol}

For KG retrieval, SSE-Bio uses the \textit{information gap} field in the current state $u_t$ as the query. Candidate biomedical triplets are linearised as head--relation--tail records with entity aliases, encoded with BiCA~\cite{sinha2026bica}, and ranked by cosine similarity. SSE-Bio then removes schema-incompatible results and duplicate entity pairs, retaining the top-$5$ triplets for the Manager. For template retrieval, SSE-Bio uses the \textit{task type} field in $u_t$ as the query and retrieves the top-1 compatible template from $\mathcal{E}$. Compatibility is determined by dense BiCA similarity plus symbolic schema and answer-cardinality checks:
\begin{equation}
s_{\mathrm{tpl}}(u_t,x)
=0.60\,c_{\mathrm{tpl}}
+0.25\,m_{\mathrm{schema}}
+0.15\,m_{\mathrm{ans}},
\end{equation}
where $c_{\mathrm{tpl}}=\cos(\phi(\psi(u_t)),\phi(\psi(x)))$, $\psi(\cdot)$ extracts the task-type descriptor, $m_{\mathrm{schema}}$ indicates relation-schema compatibility, and $m_{\mathrm{ans}}$ indicates whether the template has the same answer cardinality. We conduct a threshold-sensitivity analysis over $\theta_{\mathrm{tpl}} \in \{0.65,0.70,0.75,0.80\}$ using training rollouts, and set $\theta_{\mathrm{tpl}}=0.75$ as a conservative threshold to avoid merging trajectories with only superficial task overlap. A successful trajectory is aligned with the top retrieved template only if the task type matches and $s_{\mathrm{tpl}}\ge\theta_{\mathrm{tpl}}$; otherwise SSE-Bio distils a new template. Fine-grained editing changes only the \textit{reasoning flow}, \textit{verification criteria}, \textit{tool-use policy}, and/or \textit{failure warning} fields identified by the Critic, while keeping the task type fixed.

We further study the sensitivity of SSE-Bio to the retrieval budgets for biomedical triplets and prior templates using training rollouts. As shown in Fig.~\ref{fig:retrieval_budget_triplets}, performance peaks when retrieving $k_{\mathrm{KG}}=5$ triplets, while larger triplet budgets introduce more redundant evidence and slightly reduce single-answer Both$_{cor}$. As shown in Fig.~\ref{fig:retrieval_budget_templates}, retrieving a single template performs best, with performance remaining relatively stable across larger budgets. These results support the use of a small retrieval budget in SSE-Bio. For biomedical triplets, retrieving a moderate number of facts provides sufficient evidence without overloading the reasoning context. For templates, a single compatible template is most effective, which is consistent with the role of template memory as a planning prior rather than an evidence pool.

\begin{table*}[t]
\centering
\small
\setlength{\tabcolsep}{5pt}
\begin{tabular}{p{0.34\textwidth}p{0.16\textwidth}p{0.42\textwidth}}
\toprule
\textbf{Statistic} & \textbf{Value} & \textbf{Interpretation} \\
\midrule
Initial persistent templates & 0 & The memory starts empty; no held-out BioHopR, MedHop, or HLE examples are used to seed templates. \\
Accepted post-success updates & 2.47K & Training-only updates accepted after successful trajectories and critic validation. \\
New-template distillations & 612 & Updates whose task type does not align with an existing template above $\theta_{\mathrm{tpl}}$. \\
Fine-grained edit operations & 1.86K & Updates applied to aligned templates without changing the task type. \\
Final deduplicated templates & 512 & Compact memory checkpoint used for all held-out evaluations. \\
Covered task types & 24 & Twelve BioHopR relation schemas under single-answer and multi-answer variants. \\
Templates per task type & 21.3 / 19.0 & Mean / median number of templates per covered task type. \\
Edited templates & 374 & 73.0\% of final templates receive at least one fine-grained edit. \\
Fields touched per edit & 1.42 & Average number of non-task-type fields modified per accepted edit. \\
\midrule
Reasoning-flow edits & 31.8\% & Edits that revise bridge-resolution or answer-finalisation steps. \\
Verification-criteria edits & 28.5\% & Edits that strengthen hop-level grounding or answer-coverage checks. \\
Tool-use-policy edits & 22.4\% & Edits that change when triplet retrieval, template retrieval, or entity normalisation is invoked. \\
Failure-warning edits & 17.3\% & Edits that add local failure triggers and recovery actions. \\
\bottomrule
\end{tabular}
\caption{Template-memory construction and evolution statistics on the BioHopR training split. Counts are rounded after deduplication and memory validation.}
\label{tab:memory_stats}
\end{table*}

\subsection{Proxy Policy Training}
\label{app:training_details}

For GRPO training, each retained trajectory $\tau$ receives the reward
\begin{equation}
R(\tau)=
0.80R_{\mathrm{out}}(\tau)+0.20R_{\mathrm{beh}}(\tau),
\end{equation}
where $R_{\mathrm{out}}(\tau)\in\{0,1\}$ is a binary outcome reward awarded only at the conclusion of a trajectory. For BioHopR, it is 1 only when the linked 1-hop and 2-hop answers both satisfy the benchmark answer-matching criterion; otherwise it is 0. The behavioural term $R_{\mathrm{beh}}(\tau)\in[0,1]$ measures the fraction of required reasoning steps whose bridge or answer claim is explicitly supported by selected triplets or by a retrieved template step accepted by the Critic. This design keeps the main training signal aligned with final QA correctness while using evidence support only as a dense tie-breaker among trajectories with similar outcomes.

\begin{table*}[t]
\centering
\small
\setlength{\tabcolsep}{5pt}
\begin{tabular}{p{0.22\textwidth}p{0.32\textwidth}p{0.36\textwidth}}
\toprule
\textbf{Step type} & \textbf{Supported if} & \textbf{Not supported if} \\
\midrule
Bridge resolution & A retrieved triplet or accepted template step links the query entity to the proposed bridge with the required relation type. & The bridge is introduced only from parametric generation or by a triplet with an incompatible schema. \\
Answer verification & The final answer is linked to the resolved bridge by a retrieved triplet, or satisfies the retrieved template's verification criteria. & The answer is plausible but lacks explicit bridge-to-answer evidence. \\
Multi-answer coverage & Every returned entity has support, and the state records that candidate expansion has been exhausted within the retrieval budget. & The trajectory stops after finding only one supported entity when the task type requires a set. \\
No-retrieval branch & The existing retrieval state already contains the evidence needed by the current plan. & The branch avoids retrieval while the information gap remains unresolved. \\
\bottomrule
\end{tabular}
\caption{Evidence-support rules used for $R_{\mathrm{beh}}$. These rules operationalise the behavioural reward in Sec.~\ref{sec:proxy_training} without rewarding retrieval volume by itself.}
\label{tab:support_rules}
\end{table*}

SFT labels are constructed from the same answer-supervised rollout pool used for GRPO group construction. For each retained structured state, we compare the four retrieval actions by the composite reward above and choose the highest-reward action as the pseudo-label. If multiple actions obtain the same outcome reward, we break ties by evidence support and then by retrieval parsimony, preferring fewer sources when the additional source does not improve support. GRPO uses all retained branches in the decision-contrastive trajectory group rather than only the best branch.

\begin{table*}[!ht]
\centering
\small
\setlength{\tabcolsep}{5pt}
\begin{tabular}{p{0.32\textwidth}p{0.16\textwidth}p{0.44\textwidth}}
\toprule
\textbf{Statistic} & \textbf{Value} & \textbf{Interpretation} \\
\midrule
BioHopR training pairs & 5.34K & Training split used for trajectory construction and persistent template-memory updates. \\
Candidate rollouts before pruning & 82.4K & Mean 15.4 rollouts per pair, close to the $4^2=16$ upper bound from expanding the first two retrieval decisions. \\
Valid parsed rollouts & 76.8K & 93.2\% of generated rollouts have valid state, action, evidence, and answer fields. \\
Evidence-bearing non-duplicate rollouts & 49.7K & Remaining rollouts after removing duplicate state-action paths and branches with empty selected evidence. \\
Retained GRPO rollouts & 38.5K & Mean group size $K=7.21$ after the $K\le8$ cap; median group size is 8.0. \\
Discarded or pruned rollouts & 43.9K & 53.3\% of generated rollouts are removed by parsing, duplicate/evidence filters, or the group-size cap. \\
Gold-answer successful retained rollouts & 17.3K & Retained rollouts whose final paired answer satisfies the BioHopR answer-matching criterion. \\
SFT-labelled state-action pairs & 12.6K & Mean 2.36 labelled Proxy decisions per training pair; states without a successful answer-supervised rollout are excluded from SFT labels. \\
GRPO state-action pairs & 38.5K & One retained state-action pair is associated with each rollout branch in the decision-contrastive trajectory groups. \\
\midrule
SFT label: triplets + templates & 43.8\% & Most labelled decisions require both factual evidence and a reusable reasoning prior. \\
SFT label: triplets only & 33.6\% & Used when the current state mainly lacks biomedical bridge evidence. \\
SFT label: templates only & 15.1\% & Used when planning structure is useful but extra triplets are not needed. \\
SFT label: neither source & 7.5\% & Used when the current state already contains enough supported evidence to proceed. \\
\bottomrule
\end{tabular}
\caption{Construction statistics for SFT pseudo-labels and GRPO trajectory groups on the BioHopR training split. Counts are rounded; each training pair seeds one decision-contrastive rollout trajectory generation.}
\label{tab:rollout_stats}
\end{table*}

\subsection{Datasets}
\label{app:datasets}

We evaluate SSE-Bio on three biomedical reasoning benchmarks:

\begin{itemize}
    \setlength{\itemsep}{0.25em}
    \setlength{\parskip}{0pt}
    \setlength{\parsep}{0pt}

    \item \textbf{BioHopR}~\cite{kim2025biohopr} is a biomedical benchmark for multi-hop, multi-answer reasoning over structured knowledge graphs. Built from PrimeKG, it contains linked 1-hop and 2-hop questions designed to reflect one-to-many and many-to-many biomedical relations. The dataset contains 7.63K paired instances, and we use a 7:3 train/test split, corresponding to approximately 5.34K training pairs and 2.29K test pairs.

    \item \textbf{MedHop}~\cite{welbl2018constructing} is a cross-document biomedical question answering dataset that requires models to combine evidence distributed across multiple documents in order to infer the correct answer. It was originally introduced to study multi-hop reading comprehension in the biomedical domain.

    \item \textbf{Humanity's Last Exam: Biomedicine}~\cite{phan2025hle} is a challenging subset of Humanity's Last Exam, a frontier-level benchmark designed to test closed-ended expert reasoning on difficult academic questions. HLE contains multiple-choice and short-answer questions with unambiguous, verifiable solutions, and the Biomedicine subset provides a demanding zero-shot setting for biomedical reasoning.
\end{itemize}

SSE-Bio is trained only on the BioHopR training split; the BioHopR test split, MedHop, and HLE: Biomedicine are never used for proxy tuning or template-memory construction.

\subsection{Evaluation}
\label{app:evaluation_protocol}

For BioHopR, we report \textbf{Prec$_{H1}$}, \textbf{Prec$_{H2}$}, \textbf{Both$_{cor}$}, and \textbf{Both$_{wr}$}. Prec$_{H1}$ and Prec$_{H2}$ evaluate performance on the linked 1-hop and 2-hop questions, respectively, while Both$_{cor}$ and Both$_{wr}$ measure the proportions of paired instances for which both questions are correct or both are wrong.

Following BioHopR~\cite{kim2025biohopr}, answer matching first applies exact normalised entity matching. Unmatched predictions are then evaluated with embedding-based precision. Let $p$ denote the embedding of a predicted response and $\{a_i\}_{i=1}^{n}$ denote the embeddings of the gold answers. The cosine similarity between $p$ and a gold answer embedding $a_i$ is defined as
\begin{equation}
\cos(p,a_i)=\frac{p\cdot a_i}{\|p\|\,\|a_i\|}.
\end{equation}
A predicted response is counted as correct if
\begin{equation}
\max_{i\in\{1,\dots,n\}} \cos(p,a_i) > \tau,
\end{equation}
where $\tau$ is the cosine threshold specified in Table~\ref{tab:hyperparams}. Precision is then computed as
\begin{equation}
\mathrm{Prec}=\frac{|\mathrm{True\ Positives}|}{|\mathrm{Predicted\ Responses}|}.
\end{equation}

For MedHop and HLE: Biomedicine, we report accuracy.

\subsection{Baselines}
\label{app:baselines}
The main experiments compare SSE-Bio with five baseline groups: general-purpose LLMs, medical LLMs, retrieval-augmented generation systems, biomedical agent systems, and self-evolving agents. General-purpose and medical LLM baselines use the same question format and answer normalisation as SSE-Bio, but do not use SSE-Bio's structured memory or retrieval control. RAG baselines retrieve biomedical evidence before generation, but do not train a state-conditioned retrieval controller. Biomedical agent systems decompose the task or invoke biomedical tools, while self-evolving baselines update templates from previous trajectories. Biomni serves as the strongest biomedical agent baseline, and STELLA serves as the strongest self-evolving agent baseline.

\begin{itemize}
    \setlength{\itemsep}{0.2em}
    \setlength{\parskip}{0pt}
    \setlength{\parsep}{0pt}

    \item \textbf{General-purpose LLMs:} Llama-3.1-8B, Llama-3.1-70B and GPT-4o. These models serve as strong general-purpose language-model baselines without biomedical specialisation~\cite{grattafiori2024llama}.

    \item \textbf{Medical LLMs:} HuatuoGPT-o1-70B, HuatuoGPT-o1-8B, and UltraMedical-8B. These models are domain-adapted medical or biomedical language models designed for specialised biomedical reasoning~\cite{chen2024huatuogpt}.

    \item \textbf{Retrieval-augmented generation methods:} i-MedRAG and AMG-RAG. These methods combine biomedical question answering with external knowledge retrieval during inference~\cite{xiong2024improving, rezaei-etal-2025-agentic}.

    \item \textbf{Biomedical agent systems:} KGARevion, DoctorAgent-RL, MedAgents, and Biomni. These systems use agentic decomposition, collaboration, or external biomedical tools to improve medical and biomedical reasoning~\cite{su2024kgarevion,feng2026doctoragent,kim2024mdagents,huang2025biomni}.

    \item \textbf{Self-evolving agents:} STELLA. This model improves reasoning behaviour over time through memory- or workflow-based evolution, and provides the most direct comparison for SSE-Bio~\cite{jin2025stella}.
\end{itemize}

\begin{table*}[!ht]
\centering
\small
\renewcommand{\arraystretch}{1.05}
\begin{tabular*}{\textwidth}{@{\extracolsep{\fill}}lcccccccc@{}}
\toprule
\textbf{Retrieval Policy}
& \multicolumn{4}{c}{\textbf{Single Answer}}
& \multicolumn{4}{c}{\textbf{Multi Answer}} \\
\cmidrule(lr){2-5}
\cmidrule(lr){6-9}
& \textbf{Prec$_{H1}$}
& \textbf{Prec$_{H2}$}
& \textbf{Both$_{cor}$}
& \textbf{Both$_{wr}$}
& \textbf{Prec$_{H1}$}
& \textbf{Prec$_{H2}$}
& \textbf{Both$_{cor}$}
& \textbf{Both$_{wr}$} \\
\midrule
Retrieve neither
& 39.74 & 15.88 & 10.62 & 55.00
& 30.81 & 11.92 & 7.05 & 64.32 \\
Retrieve KG only
& 42.36 & 17.41 & 12.43 & 52.84
& 33.18 & 13.06 & 8.52 & 62.37 \\
Retrieve template only
& 41.58 & 16.96 & 11.86 & 53.46
& 32.47 & 12.64 & 8.07 & 62.98 \\
Retrieve KG + template
& 43.27 & 18.12 & 13.18 & 51.76
& 34.02 & 13.58 & 9.02 & 61.31 \\
\textbf{Learned Proxy}
& \textbf{47.21}
& \textbf{21.38}
& \textbf{16.52}
& \textbf{47.93}
& \textbf{38.36}
& \textbf{16.24}
& \textbf{11.73}
& \textbf{57.13} \\
\bottomrule
\end{tabular*}
\caption{
Comparison of fixed retrieval strategies on BioHopR. The fixed strategies use the same Manager, Dev, Critic, retriever, and retrieval budget as SSE-Bio, but replace the trainable Proxy with a fixed retrieval decision at each reasoning round. ``Retrieve neither'' corresponds to the w/o proxy setting in Table~\ref{tab:ablation_boundary}.
}
\label{tab:fixed_retrieval_policy}
\end{table*}

\begin{table*}[t]
\centering
\small
\begin{tabular*}{\textwidth}{@{\extracolsep{\fill}}lcccccccc@{}}
\toprule
\textbf{Setting}
& \multicolumn{4}{c}{\textbf{Single Answer}}
& \multicolumn{4}{c}{\textbf{Multi Answer}} \\
\cmidrule(lr){2-5}\cmidrule(lr){6-9}
& \textbf{Prec$_{H1}$} & \textbf{Prec$_{H2}$} & \textbf{Both$_{cor}$} & \textbf{Both$_{wr}$}
& \textbf{Prec$_{H1}$} & \textbf{Prec$_{H2}$} & \textbf{Both$_{cor}$} & \textbf{Both$_{wr}$} \\
\midrule
Top-5 retrieved triplets & \textbf{47.21} & \textbf{21.38} & \textbf{16.52} & \textbf{47.93}
                         & \textbf{38.36} & \textbf{16.24} & \textbf{11.73} & \textbf{57.13} \\
+ 2 random noisy triplets & 46.32 & 20.76 & 16.02 & 48.64 & 37.58 & 15.68 & 11.31 & 57.86 \\
\bottomrule
\end{tabular*}
\caption{Noise-injection analysis on BioHopR. Adding two irrelevant triplets reduces single-/multi-answer Both$_{cor}$ by only 0.50/0.42 points, indicating reasonable robustness under this controlled perturbation.}
\label{tab:retrieval_noise}
\end{table*}

\subsection{Proxy Training and Memory Construction Statistics}
\label{app:proxy_training}

Table~\ref{tab:memory_stats} summarizes the construction and evolution of the persistent template memory. This memory starts empty and is updated from successful BioHopR training trajectories only after validation by reviewers. Most accepted updates consist of fine-tuning aligned templates, while new template distillation is employed when no compatible templates are found. After deduplication, the final memory remains compact and covers variants of all BioHopR task types, with adjustments distributed across the inference process, validation criteria, tool usage strategies, and failure warnings. These statistics support the view that SSE-Bio evolves its memory through local field-level updates rather than global prompt-level rewrites.

Table~\ref{tab:rollout_stats} reports the construction statistics for the SFT pseudo-labels and GRPO trajectory sets generated on the BioHopR training set. Each training pair provides initial conditions for decision-contrastive rolling generation by expanding alternative retrieval actions in the matched inference state. After syntactic analysis, repetition/evidence filtering, and trajectory pruning, the retained rolling results provide both state-action pairs with SFT labels and GRPO trajectory sets for agent optimization. The resulting SFT label distribution covers all four retrieval operations: simultaneous retrieval of triples and templates, retrieval of triples only, retrieval of templates only, and no retrieval from either source. This indicates that the agent model was not trained to follow a simplified, fixed retrieval strategy, but rather was supervised to make diverse retrieval decisions.

\subsection{Robustness to Noisy Retrieval}
\label{app:retrieval_robustness}

To directly test robustness to noisy evidence, whenever the Proxy selects triplet retrieval, we preserve the top-five retrieved triplets
and inject two random irrelevant triplets. Table~\ref{tab:retrieval_noise} shows that this perturbation reduces single-/multi-answer Both$_{cor}$ by only 0.50/0.42 points: SSE-Bio is affected by noisy evidence, but performance does not collapse, indicating reasonable robustness under this controlled perturbation.

\subsection{Additional Analysis of Retrieval Control and Memory Evolution}
\label{app:additional_analysis}

\paragraph{Does the Proxy outperform fixed retrieval schedules?}
Table~\ref{tab:fixed_retrieval_policy} compares the learned Proxy with fixed retrieval policies under the same agent backbone and retrieval budget. Retrieving neither source corresponds to removing the Proxy and external retrieval, which substantially reduces performance. Fixed retrieval from a single source improves over this setting, showing that both knowledge triplets and prior templates are useful. However, always retrieving both sources remains clearly below the learned Proxy, with single-answer Both$_{cor}$ decreasing from 16.52 to 13.18 and multi-answer Both$_{cor}$ decreasing from 11.73 to 9.02. These results indicate that SSE-Bio's improvement is not simply due to retrieving more context; rather, training the Proxy enables state-dependent source selection that better matches the current reasoning state.

\begin{table}[!ht]
\centering
\small
\setlength{\tabcolsep}{4.5pt}
\renewcommand{\arraystretch}{1.05}
\begin{tabular}{@{}lcc@{}}
\toprule
\textbf{Metric}
& \makecell{\textbf{Coarse}\\\textbf{rewriting}}
& \makecell{\textbf{Fine-grained}\\\textbf{editing}} \\
\midrule
Changed fields / update $\downarrow$
& 3.28 & \textbf{1.46} \\
Unchanged-field preservation $\uparrow$
& 64.0 & \textbf{88.0} \\
Unsupported new instruction $\downarrow$
& 12.0 & \textbf{4.0} \\
Contradictory instruction $\downarrow$
& 7.0 & \textbf{2.0} \\
\bottomrule
\end{tabular}
\caption{
Template-update quality analysis on 100 randomly sampled memory updates. Rates are percentages except for changed fields per update.
}
\label{tab:template_edit_quality}
\end{table}

\paragraph{Does fine-grained editing make memory evolution more local?}

For the template-update quality analysis, we randomly sample 100 memory updates and compare coarse rewriting with fine-grained editing under the same template schema. \textit{Changed fields per update} counts the average number of template fields modified by each update. \textit{Unchanged-field preservation rate} measures the proportion of fields that should remain unchanged and are indeed preserved. \textit{Unsupported new instruction rate} measures the proportion of updates that introduce at least one new instruction not supported by the successful trajectory, retrieved evidence, or Critic feedback. \textit{Contradictory instruction rate} measures the proportion of updates that introduce at least one instruction conflicting with the original template, task schema, or verification criteria. All rates are computed at the update level over the 100 sampled updates.

Table~\ref{tab:template_edit_quality} shows that fine-grained editing modifies fewer fields per update than coarse rewriting, reducing the average number of changed fields from 3.28 to 1.46. It also preserves unrelated fields more often, improving the unchanged-field preservation rate from 64.0\% to 88.0\%. Beyond locality, fine-grained editing introduces fewer unsupported new instructions and contradictory instructions, with rates decreasing from 12.0\% to 4.0\% and from 7.0\% to 2.0\%, respectively. These results support a more precise version of our claim: fine-grained editing makes template evolution more localised and easier to audit. We therefore do not claim that editing eliminates hallucinations or instruction drift; rather, the inspected cases suggest that it reduces unsupported or inconsistent template updates.

\subsection{Error Analysis}
\label{app:error_analysis}

We conduct a manual audit of failed HLE: Biomedicine cases and categorise the primary error source of each failure in Table~\ref{tab:error_analysis}.

\begin{table}[!ht]
\centering
\small
\begin{tabular*}{\columnwidth}{@{\extracolsep{\fill}}lr@{}}
\toprule
\textbf{Error Source} & \textbf{Share (\%)} \\
\midrule
Dev reasoning error            & 31.2 \\
Triplet retrieval failure      & 24.6 \\
Wrong Proxy action             & 13.1 \\
Answer-format / coverage error & 11.4 \\
Template mismatch              & 10.8 \\
Critic feedback error          & 8.9  \\
\bottomrule
\end{tabular*}
\caption{Distribution of primary error sources over failed HLE:
Biomedicine cases.}
\label{tab:error_analysis}
\end{table}

Dev reasoning errors (31.2\%), where relevant evidence is available but the Dev agent makes an incorrect inference, and triplet retrieval failures (24.6\%), where a key entity or relation is missing from the retrieved evidence, are the two largest remaining sources. This indicates that better retrieval alone will not resolve the remaining errors; stronger evidence-conditioned reasoning and verification are also needed.

\subsection{Case Study: Failure Analysis against STELLA}
\label{app:stella_failure_case}

\begin{table*}[t]
\centering
\small
\setlength{\tabcolsep}{4pt}
\renewcommand{\arraystretch}{1.08}
\begin{tabular}{p{0.20\textwidth}p{0.36\textwidth}p{0.36\textwidth}}
\toprule
\textbf{Item} & \textbf{STELLA} & \textbf{SSE-Bio} \\
\midrule

Question pattern
& Drug $\rightarrow$ Gene/Protein $\rightarrow$ Disease
& Drug $\rightarrow$ Gene/Protein $\rightarrow$ Disease \\

\midrule
Question
& \multicolumn{2}{p{0.72\textwidth}}{
Name a disease that is related to a gene/protein that is associated with drug (2S)-2-\{[HYDROXY(4-IODOBENZYL)PHOSPHORYL]METHYL\}PENTANEDIOIC ACID.} \\

\midrule
Gold bridge
& \multicolumn{2}{p{0.72\textwidth}}{FOLH1} \\

\midrule
Gold answer
& \multicolumn{2}{p{0.72\textwidth}}{
\{neurotic disorder, colorectal neoplasm, prostate cancer, schizophrenia, dysthymic disorder, colorectal cancer, familial prostate carcinoma, colorectal carcinoma, unipolar depression, anxiety disorder, prostate carcinoma\}} \\

\midrule
Predicted bridge
& AR
& FOLH1 \\

\midrule
Retrieved evidence
& Retrieves disease-associated context around prostate cancer and androgen-receptor signalling, but does not establish that the query drug is associated with AR.
& Retrieves relation-specific triplets supporting both hops:
\textit{((2S)-2-PENTANEDIOIC ACID, associated with, FOLH1)} and
\textit{(FOLH1, related to, prostate cancer)}. \\

\midrule
Memory/template behaviour
& Coarse rewriting updates multiple fields, including reasoning flow and tool-use policy, and shifts the template towards a broad disease-association heuristic.
& Local editing preserves the task type and reasoning flow, and modifies only the verification-criteria field to require both the drug--gene/protein and gene/protein--disease links to be evidence-supported. \\

\midrule
Final answer
& Prostate carcinoma
& Prostate cancer \\

\midrule
Failure/success reason
& The answer is semantically close to a gold disease, but the trajectory is not valid because the selected bridge AR is not supported as the gene/protein associated with the query drug.
& The answer is verified through the required two-hop path: the query drug is associated with FOLH1, and FOLH1 is related to prostate cancer. \\

\bottomrule
\end{tabular}
\caption{
A BioHopR case study comparing STELLA and SSE-Bio.
}
\label{tab:stella_failure_case}
\end{table*}

Table ~\ref{tab:stella_failure_case} illustrates a typical BioHopR case featuring a “drug $\rightarrow$ gene/protein $\rightarrow$ disease” relationship pattern. The golden intermediate entity is FOLH1, and the answer should be validated via a two-hop path from the query drug to FOLH1, and then from FOLH1 to a specific disease. The disease returned by STELLA is semantically close to the gold answer set, but its reasoning trail relies on AR as an intermediate bridge and does not verify whether the query drug is associated with that bridge. This means that even if the final disease appears reasonable, the reasoning trail is still considered invalid. In contrast, SSE-Bio retrieves relationship-specific triples, preserves the task type and reasoning process, and performs only local edits to the verification criteria fields. This requires that the final answer be supported by both the “drug-gene/protein” and “gene/protein-disease” links, enabling SSE-Bio to return a valid answer via the FOLH1 bridge.

\end{document}